%% file: unsupervised2020.tex
\documentclass[11pt,a4paper]{article}
\usepackage[hyperref]{acl2021}
\usepackage{times}
\usepackage{latexsym}

\usepackage{microtype}

\aclfinalcopy % Uncomment this line for the final submission
\author{Jonas Groschwitz \\
  Saarland University\\
  \texttt{ jonasg@coli.uni-saarland.de} \\\And
  Meaghan Fowlie \\
  Utrecht University \\
  \texttt{m.fowlie@uu.nl} \\\AND
  Alexander Koller \\
  Saarland University\\
  \texttt{ koller@coli.uni-saarland.de}
  }

\date{}

\include{macros}

\title{Learning compositional structures for semantic graph parsing}

\date{}

\begin{document}
\maketitle

\begin{abstract}

AM dependency parsing is a method for neural semantic graph parsing that exploits the principle of compositionality. While AM dependency parsers have been shown to be fast and accurate across several graphbanks, they require explicit annotations of the compositional tree structures for training. In the past, these were obtained using complex graphbank-specific heuristics written by experts. Here we show how they can instead be trained directly on the graphs with a neural latent-variable model, drastically reducing the amount and complexity of manual heuristics. We demonstrate that our model picks up on several linguistic phenomena on its own and achieves comparable accuracy to supervised training, greatly facilitating the use of AM dependency parsing for new sembanks.
 
\end{abstract}

\input introduction
\input{relwork}

\input background
\input decomposition-new
\input automata

\input neural

\input results
\input{linguistic-analysis}

\section{Conclusion}
In this work, we presented a method to obtain the compositional structures for AM dependency parsing that relies much less on graphbank-specific heuristics written by experts. Our neural model learns linguistically meaningful argument slot names, as shown by our manual evaluation; in this regard, our model learns to do the job of the linguist. High parsing performance across graphbanks shows that the learned compositional structures are also well-suited for practical applications, promising easier adaptation of AM dependency parsing to new graphbanks.

% \todo{future work?}

% \todo{MOVE APPENDIX TO SEPARATE FILE before submission}

\section*{Acknowledgments}
We would like to thank the anonymous reviewers as well as Lucia Donatelli, Pia Wei\ss{}enhorn and Matthias Lindemann for their thoughtful comments.
This research was in part funded by the Deutsche Forschungsgemeinschaft (DFG, German Research Foundation), project KO 2916/2-2, and by the Dutch Research Council (NWO) as part of the project \textit{Learning Meaning from Structure} (VI.Veni.194.057).

\clearpage

\bibliographystyle{acl_natbib}
\bibliography{mybib,emnlp2020}

\clearpage
\appendix

% sampling method for hand analysis
\input{sampling}
\input{appendix-all-trees-alg}
\input{other-details}

\clearpage

\end{document}

%% file: macros.tex
\usepackage{microtype}

\usepackage{qtree}
\usepackage{forest}

\usepackage{float}
\usepackage{placeins}
\usepackage{graphicx} % for \rotatebox
\usepackage{amsmath}
\usepackage{amsthm}
\usepackage{amssymb}
\usepackage{latexsym}
\usepackage{mathtools}

\newtheorem{theorem}{Theorem}

\usepackage{algpseudocode}%[noend]
\usepackage[ruled,vlined, linesnumbered]{algorithm2e}

\usepackage{booktabs}

\usepackage{caption}
\usepackage{subcaption}

\usepackage{bussproofs} 
\usepackage{stackengine}

\usepackage{enumerate}

\newcommand{\ra}{\ensuremath{\rangle}\xspace}
\newcommand{\la}{\ensuremath{\langle}}

\newcommand{\N}{\ensuremath{\mathbb{N}}}
\newcommand{\nul}{\ensuremath{\emptyset}}
\newcommand{\sortof}[1]{`#1'}
\newcommand{\of}[1]{\left(#1\right)}

\newlength\mylen

\makeatletter
\def\th@definition{%
  \thm@notefont{}% same as heading font
  \normalfont % body font
}
\makeatother

\theoremstyle{definition} % make the font normal, not italics
\theoremstyle{plain} % these should be in italics
\newtheorem{thm}{Theorem}[section]
\newtheorem{lemma}[thm]{Lemma}

\theoremstyle{definition}

\usepackage{amssymb}

\usepackage{pifont}

\usepackage{setspace}

\usepackage{bm}
\usepackage{xspace}
\usepackage[utf8]{inputenc}

\newcommand{\sources}{\ensuremath{\mathcal{S}}}
\newcommand{\type}[1]{\ensuremath{[#1]}}

\newcommand{\app}[1]{\text{\textsc{App}}\ensuremath{_{#1}}}  %\app{arg}
\newcommand{\modify}[1]{\text{\textsc{Mod}}\ensuremath{_{#1}}\xspace}

\newcommand{\src}[1]{\text{\textsc{#1}}\xspace}                                  % source name 
\newcommand{\obj}{\text{\src{o}}}               % object sources. \obj or \obj[i]
\newcommand{\subj}{\text{\src{s}}}													% subject source, just \subj
\newcommand{\rt}{\ensuremath{r}}
\newcommand{\modsrc}{\text{\src{m}}}

\newcommand{\G}[1]{\text{\textit{G-#1}}} % naming graphs with real sources
\renewcommand{\H}[1]{\text{\textit{P-#1}}} % naming canonical constants
\newcommand{\Hhat}[1]{\text{\textit{P\ensuremath{'}-#1}}} % naming constants with placeholder sources
\newcommand{\emptytype}{\ensuremath{[\,]}}

\newcommand{\el}[1]{\text{`#1'}}

 \newcommand{\oper}{\ensuremath{\ell}}  % operation (edge label)
\newcommand{\edge}[3]{\ensuremath{#1\xrightarrow{\text{\tiny #3}}#2}}% edge from #1 to #2 with label #3

\newcommand{\score}{\cost} % just for legacy
\newcommand{\cost}{\ensuremath{c}}
\newcommand{\edgeCost}[3]{\ensuremath{\cost\of{\edge{#1}{#2}{#3}}}} % cost of edge from #1 to #2 with label #3
\newcommand{\graphCost}[2]{\ensuremath{\cost\of{#2,#1}}} % cost of supertag #2 at word #1

\usepackage{tikz}
\usepackage{tikz-qtree}

\usetikzlibrary{graphs}
\usetikzlibrary{shapes,arrows}
\usetikzlibrary{positioning}
\usetikzlibrary{quotes}

\tikzset{snode/.style={
    ellipse,
    minimum size=6mm,
    very thick,
    draw=black,
    font=\rmfamily}}

\tikzset{ssrc/.style={
    ellipse,
    minimum size=6mm,
    very thick,
    fill=black!10,
    draw=black,
    text=black,
    font=\rmfamily}}

\tikzset{sanno/.style={node distance=-1mm,font=\rmfamily\small}}

\tikzset{sedgel/.style={
    color=black,
    sloped, above,
    font=\rmfamily\small}}

\tikzset{sedge/.style={
    thick,
    >=stealth'
}}

\usepackage{array}

\newcommand{\cin}[1]{\scalebox{0.8}{\ensuremath{\pm #1}}}

\usepackage{color, colortbl}
\usepackage{bm}
\definecolor{orange}{rgb}{1,0.5,0}
\definecolor{mdgreen}{rgb}{0.05,0.6,0.05}
\definecolor{mdblue}{rgb}{0,0,0.7}
\definecolor{dkblue}{rgb}{0,0,0.5}
\definecolor{dkgray}{rgb}{0.3,0.3,0.3}
\definecolor{dkgreen}{rgb}{0,0.5,0}
\definecolor{slate}{rgb}{0.25,0.25,0.4}
\definecolor{gray}{rgb}{0.5,0.5,0.5}
\definecolor{ltgray}{rgb}{0.7,0.7,0.7}
\definecolor{purple}{rgb}{0.7,0,1.0}
\definecolor{lavender}{rgb}{0.65,0.55,1.0}
\definecolor{brown}{rgb}{0.6,0.2,0.2}

\newcommand{\ignore}[1]{}

\newcommand{\sent}[1]{\emph{#1}}

\newcommand{\lastACL}{L'19}

\newcommand{\myref}[1]{\S\ref{#1}}

\newcommand{\nn}[1]{\ensuremath{#1}}  % node name in text
\newcommand{\nntree}[1]{\text{#1}}  % node name in trees

\newcommand{\ps}[1]{\ensuremath{\mathsf{#1}}} % Options: \mathbbmss or \mathbbm from package bbm, \mathcal from boondox-cal or boondox-calo, \mathfrak from amssymb
\newcommand{\refnode}[1]{\ensuremath{\text{REF-}#1}}
\newcommand{\resolutionTarget}[1]{\ensuremath{\text{RT}\of{#1}}}
\newcommand{\termtype}[1]{\ensuremath{\beta}(#1)}
\newcommand{\lexicaltype}[1]{\ensuremath{\tau}(#1)}

\newcommand{\depNodeWType}[2]{{#1}:#2}
\newcommand{\depNodeWTypeNode}[3]{\nntree{#3}:{#1}:#2}

\newcommand{\arule}[3]{\ensuremath{#2\of{#3}\rightarrow#1}}

\newcommand{\astate}[1]{\ensuremath{\left\langle#1\right\rangle}}

\newcommand{\defAMRScale}{0.5}

\usepackage{forest}
\usepackage{tikz-dependency}
\depstyle{amdep}{label style={text=black, fill=black!10}, edge style={thick}, edge slant=1pt, label style={font=\large}}
\depstyle{syntaxdep}{text only label, label style={font=\small\slshape, above}, edge style={>=latex,solid,gray}, edge slant=5pt}

\usepackage{linguex}

%% file: introduction.tex
\section{Introduction} \label{sec:introduction}

It is generally accepted in linguistic semantics that meaning is
\emph{compositional}, i.e.\ that the meaning representation for a
sentence can be computed by evaluating a tree bottom-up. 
A compositional parsing model not only reflects this insight, but has practical advantages such as in compositional generalisation (e.g.~\citealt{herzig2020spanbased}), i.e.~systematically generalizing from limited data.

However, in developing a compositional
semantic parser, one faces the task of figuring out 
what exactly the compositional structures -- i.e.~the trees that
link the sentence and the meaning representation -- should look like.
% how to
% decompose the target meaning representation for a sentence into the
% pieces that can then be compositionally recombined.
This is
challenging even for expert linguists; for instance, \cite{copestake2001algebra}
report that 90\% of the development time of the English Resource
Grammar \cite{copestake00:_englis_hpsg} went into the development of
the syntax-semantics interface.

Compositional semantic parsers which are learned from data face an
analogous problem: to train a such a parser, the compositional structures must be made explicit. However, these structures are not annotated in most sembanks.
For instance, the \emph{AM (Apply-Modify) dependency parser} of
\newcite{groschwitz18:_amr_depen_parsin_typed_seman_algeb} uses a
neural model to predict \emph{AM dependency trees}, compositional structures that evaluate to semantic graphs.
Their parser achieves
high accuracy \cite{lindemann-etal-2019-compositional} and parsing
speed \cite{lindemann-etal-2020-fast} across a variety of English semantic
graphbanks. 
%, including the AMRBank \cite{amBanarescuBCGGHKKPS13} and the Semantic Dependency Parsing graphbanks \cite{OepenKMZCFHU15}.
To obtain an AM dependency tree for each graph in the corpus, they use hand-written
graphbank-specific heuristics.
%for decomposing the annotated graphs
%into AM dependency trees.
These heuristics cost significant time and
expert knowledge to create, limiting the ability of the AM parser
to scale to new sembanks.

In this paper, we drastically reduce the need for hand-written heuristics
for training the AM dependency parser.
% In this paper, we show how to automatically decompose semantic graphs
% into AM dependency trees and train the AM dependency parser without
% requiring hand-written heuristics.
We first present a graphbank-independent method to compactly represent the relevant
compositional structures of a graph in a tree automaton.
% We offer graphbank-independent theoretical results that strongly constrain
% the compositional structure of a graph and show how
% to compactly represent \todo{the} remaining options in a tree automaton.
We then train a neural AM dependency parser directly
on these tree automata.
%then demonstrate how the AM dependency parser can be trained directly
%on these tree automata.
%, using either an EM algorithm based on
%PCFG-style rule weights or using gradient descent for a neural
%model.
Our code is available at \url{github.com/coli-saar/am-parser}.

We evaluate the consistency and usefulness of the learned
compositional structures in two ways. We first evaluate the accuracy
of the trained AM dependency parsers, across four graphbanks, and find
that it is on par with an AM dependency parser that was trained on the
hand-designed compositional structures of
\newcite{lindemann-etal-2019-compositional}. We then analyze the
compositional structures which our algorithm produced, and find that
they are linguistically consistent and meaningful. We expect that our
methods will facilitate the design of compositional models of
semantics in the future.

% \begin{enumerate}
%     \item Based on linguistic principles.
%     \item The abstraction to use compositional structure allows using the same parsing method for different graphbanks.
%     \item Strong performance.
% \end{enumerate}

% However, it has a disadvantage: the compositional structure is
% latent; so far it has been determined with handwritten heuristics
% \citep{lindemann-etal-2019-compositional}. Here we develop a new
% method, based on tree automata, to learn the compositional structure
% during training of the parser.

%%% Local Variables:
%%% mode: latex
%%% TeX-master: "unsupervised2020"
%%% End:

%% file: relwork.tex
\section{Related work}\label{sec:relwork}

\input{background-figure} % not sure why this needs to be here, but it gives the desired result

Compositional semantic graph parsers other than AM dependency parsers, like \newcite{ArtziLZ15}, \newcite{PengSG15} and \newcite{chen-etal-2018-accurate}, use CCG and HRG based grammars to parse AMR and EDS \citep{OpenSDP}. They use a combination of heuristics, hand-annotated compositional structures and sampling to obtain training data for their parsers, in contrast to our joint neural technique. None of these approaches use slot names that carry meaning; to the best of our knowledge this work is the first to learn them from data.

% Other formalisms beside AM dependency parsing have been used for compositinal semantic graph parsing. \newcite{ArtziLZ15} use a CCG-based method for AMR parsing; they insert expert knowledge into their system by obtaining a seed lexicon from graphs hand-annotated with compositional structures. \cite{PengSG15} learn Hyperedge-Replacement Grammars (HRG) for AMR parsing; they use heuristics similar to the blob-heuristics we use here to attach edges to nodes to reduce the latent variable space and they perform MCMC sampling to learn the rest of the grammar. \cite{chen-etal-2018-accurate} apply HRG to EDS parsing, relying on the direct connection of EDS to the expert-developed English Resource Grammar (ERG) and the syntactic annotation of DeepBank to extract their grammar. None of these approaches use slot names that carry meaning; to the best of our knowledge this work is the first to learn them from data.

\newcite{fancellu-etal-2019-semantic} use DAG grammars for compositional parsing of Discourse Representation Structures (DRS). Their algorithm for extracting the compositional structure of a graph is deterministic and graphbank-independent, but comes at a cost: for example, rules for heads require different versions depending on how often the head is modified, reducing the reusability of the rule.
%, and reentrancies cause not only meaningful rule variations like the type annotations of the AM dep-trees, but also somewhat spurious variations for the nodes where the reentrancy occurs. This has the potential to hinder how well the rules can generalize.

\newcite{maillard2019jointly} and \newcite{havrylov-etal-2019-cooperative} learn compositional, continuous-space neural sentence encodings using latent tree structures.  %These methods use unrestricted charts, as opposed to the type-restricted tree automaton we use here; i.e.~they operate in a structurally less complex setting. Further, 
Their tasks are different: they learn to predict continous-space embeddings; we learn to predict symbolic compositional structures.
Similar observations hold for self-attention \citep{vaswani2017attention, kitaev-klein-2018-constituency}.

% Learning compositional structures plays a role beyond explicitly symbolic compositional parsers. When obtaining vector representations for a sentence, methods such as self-attention \citep{vaswani2017attention, kitaev-klein-2018-constituency} or backpropagation through a CKY chart \citep{maillard2019jointly} learn implicit, continuous constituency structures that guide the neural encoder. Neural module networks \citep{andreas2016neural} learn more discrete ways of composing neural modules into an encoder.

% \begin{itemize}
%     \item Other unification-style formalisms that use slot names. HPSG? HR algebra? Anything else? Did anyone learn names for them?
%     \item Kitaev \& Klein \url{https://arxiv.org/pdf/1805.01052.pdf} and Maillard et al \url{https://arxiv.org/pdf/1705.09189.pdf} learn latent compositional structures, or rather a real-valued distribution of them, to structure their sentence encoders. We by contrast learn discrete compositional structures that evaluate to graphs. Also, for us, the tree structure is fixed and we learn the source names instead.
%     \item Three currently competitive paradigms of semantic graph parsing: seq2seq, graph-based, compositional. By making the compositional structure explicit, compositional graph parsers expose some of their inner workings to human eyes and offer the promise of a more appropriate inductive bias.
%     \item How do \cite{ArtziLZ15} and \cite{PengSG15} and \cite{chen-etal-2018-accurate} get their grammars? I think heuristics and some EM-like things, but not sure, need to read.
% \end{itemize}

%% file: background-figure.tex
\begin{figure*}[t!]
\centering
\begin{minipage}{0.71\linewidth}

% Aligned AM tree
\begin{subfigure}[b]{0.45\linewidth}
    \pgfkeys{/pgf/inner sep=0.1em} % words close together
    \begin{forest}  
    sn edges/.style={for tree={
    align=center, font=\small, l sep=0.5em, l=0.5em,
    parent anchor=south, child anchor=north}},  
    sn edges,
    [\G{fairy}
        [,phantom
            [The, tier=word]
        ]
        [fairy, tier=word, edge=dashed]
        [\G{begin}, edge=->, edge label = {node [midway, right, xshift=.4em, yshift=.25em] {\small\modify{\subj}}}, l=1.4em
        [,phantom
            [that, tier=word]
        ]
            [begins,tier=word, edge=dashed]
            [,phantom
                [to, tier=word]
            ]
            [\G{glow}, edge=->, edge label = {node [midway, right, xshift=.3em, yshift=.25em] {\small\app{\obj}}}, l=1.7em
                [glow,tier=word, edge=dashed]
            ]
        ]
    ]
    \end{forest}
    \caption{AM dep-tree with word alignments. The dashed lines connect tokens to their graph constants, and arrows point from heads to arguments, labeled by the operation that puts the graphs together.}
    \label{fig:am-tree}
    \end{subfigure}~~
    % Unaligned AM tree
    \begin{subfigure}[b]{0.29\linewidth}
    \centering
    \begin{forest}
    for tree={align=center, edge=->, font=\small, l sep=1.2em, l=1.2em, }
    [\depNodeWType{\G{fairy}}{\emptytype}
        [\depNodeWType{\G{begin}}{\type{\subj{}, \obj\type{\subj}}}, edge label = {node [midway, right] {\small\modify{\subj}}}
            [\depNodeWType{\G{glow}}{\type{\subj}}, edge label = {node [midway, right] {\small\app{\obj}}}
            ]
        ]
    ]
    \end{forest}
    \caption{AM dep-tree without alignments. Nodes are labeled with graph constants, paired with their types for ease of presentation.}
    \label{fig:relative-unaligned}
    \end{subfigure}~~
    \begin{minipage}[b]{0.21\linewidth}
        % AMR
        \begin{subfigure}[b]{\linewidth}
            \includegraphics[scale=\defAMRScale]{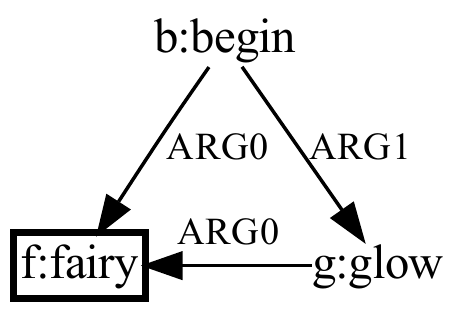}
            \caption{AMR}
            \label{fig:amr-fairy}
        \end{subfigure}\\[12pt]
        % partial result
        \begin{subfigure}[b]{\linewidth}
            \centering
            \includegraphics[scale=\defAMRScale]{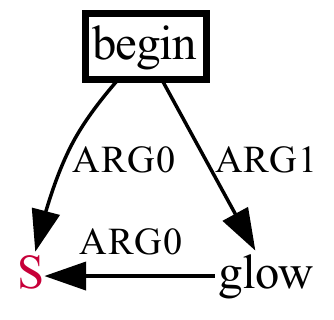}
            % \caption{Partial result: \textit{begins to glow}}
            \caption{Partial result: \textit{begins to glow}}
            \label{fig:partial-VP}
        \end{subfigure}
    \end{minipage}

\caption{AM dep-trees and graphs for \textit{the fairy that begins to glow}. We usually write our example AM dep-trees without alignments as in (b). We include node names where helpful, as in (c), where e.g.\ \nn{b} is labeled \sent{begin}.}
    \vspace{-5pt}
\label{fig:am-term}
\end{minipage}\quad~
\begin{minipage}{0.25\linewidth}
\begin{minipage}{\linewidth}
\centering
\begin{subfigure}[t]{0.35\linewidth}
    \centering
    \includegraphics[scale=\defAMRScale]{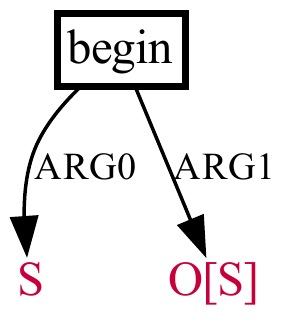}
    \caption*{\G{begin}}
    \label{fig:begin}
\end{subfigure}
\begin{minipage}[b]{0.25\linewidth}
\begin{subfigure}[t]{\linewidth}
\centering
    \includegraphics[scale=\defAMRScale]{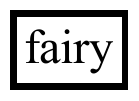}
    \caption*{\G{fairy}}
    \label{fig:fairy}
\end{subfigure}\\[12pt]
\begin{subfigure}[t]{\linewidth}
\centering
    \includegraphics[scale=\defAMRScale]{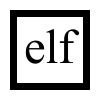}
    \caption*{\G{elf}}
    \label{fig:elf}
\end{subfigure}
\end{minipage}
\begin{subfigure}[t]{0.25\linewidth}
    \centering
    \includegraphics[scale=\defAMRScale]{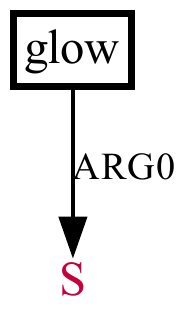}
    \caption*{\G{glow}}
    \label{fig:glow}
\end{subfigure}
% \begin{subfigure}[t]{0.11\linewidth}
%     \includegraphics[scale=\defAMRScale]{pics/tiny_f.pdf}
%     \caption*{\G{tiny}}
%     \label{fig:tiny}
% \end{subfigure}
\end{minipage}\vspace{11pt}
\begin{minipage}{\linewidth}
\centering
\begin{subfigure}[t]{0.48\linewidth}
\centering
    \includegraphics[scale=\defAMRScale]{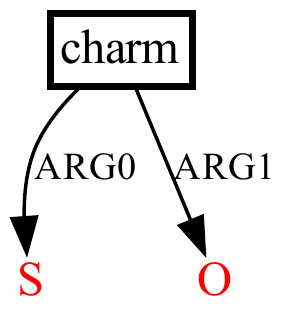}
    \caption*{\G{charm}}
    \label{fig:charm-active}
\end{subfigure}
\begin{subfigure}[t]{0.48\linewidth}
\centering
    \includegraphics[scale=\defAMRScale]{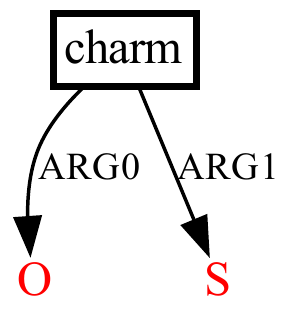}
    \caption*{\G{charmP}}
    \label{fig:charm-passive}
\end{subfigure}
\caption{Graph constants}
\label{fig:as-graphs}
\end{minipage}
\end{minipage}\vspace{-5pt}

\end{figure*}

%% file: background.tex
\input{constantsFigure} % again, not sure why this needs to be so far up

\section{AM dependency parsing}\label{sec:background}

Compositional semantic graph parsing methods do not predict a graph directly, but rather predict a compositional structure which in turn determines the graph.
% In AM dependency parsing, the tree in Fig.\ \ref{fig:am-term} describes the way the meanings of the words -- graph fragments in Fig.\ \ref{fig:as-graphs} -- combine to form the AMR in Fig.\ \ref{fig:amr-fairy}.
% representing the meaning of the sentence: \textit{begins} combines with \textit{glow}; meanwhile
%\textit{fairy} combines with its modifier \textit{tiny}, yielding a graph that forms the subject of \textit{begins}. How exactly they combine depends on the operations annotated on the tree edges, which are explained below (\S\ref{sec:am-depend-pars}). 
\citet{groschwitz18:_amr_depen_parsin_typed_seman_algeb} represent the compositional structure of a graph with \textbf{\emph{AM dependency trees}} (AM dep-trees for short) like the one in Fig.\ \ref{fig:am-tree}. It describes the way the meanings of the words -- the graph fragments in Fig.~\ref{fig:as-graphs} -- combine to form the semantic graph in Fig.~\ref{fig:amr-fairy}, here an AMR \citep{amBanarescuBCGGHKKPS13}. The AM dep-tree edges are labeled with graph-combining operations, taken from the \emph{Apply-Modify (AM) algebra} \citep{groschwitz-etal-2017-constrained,GroschwitzDiss}. 

% A supertagger assigns a \emph{graph constant} to each token representing its lexical meaning. An edge-factored model predicts which graph-building operations combine which graph constants;

% \citet{groschwitz18:_amr_depen_parsin_typed_seman_algeb} use \emph{AM
%   dependency trees} to represent the compositional structure of a
% semantic graph. Each token is assigned a \emph{graph constant}
% representing its lexical meaning; dependency labels correspond to
% operations of the \emph{Apply-Modify
%   (AM) algebra} \citep{groschwitz-etal-2017-constrained,GroschwitzDiss}, which
% combine graphs into bigger ones.

Graphs are built out of fragments called \textbf{\textit{graph constants}}
(Fig.~\ref{fig:as-graphs}). Each graph constant has a \textbf{\emph{root}}, marked with a rectangular outline, and may have special node markers called \textbf{\emph{sources}} \citep{CourcelleE12}, drawn in red, which mark the empty slots where other graphs will be inserted.

In Fig.~\ref{fig:am-tree}, the \app{\obj} operation plugs the root of \G{glow} into the \obj{} source of \G{begin}. Because \G{begin} and \G{glow} both have an \subj{}-source, \app{\obj} merges these nodes, creating a \textbf{\textit{reentrancy}}, i.e.\ an undirected cycle, and yielding Fig.~\ref{fig:partial-VP}, which is in turn attached at \subj{} to the root of \G{fairy} by \modify{\subj{}}. \app{} fills a source of a head with an argument while \modify{} uses a source of a modifier to connect it to a head; both operations keep the root of the head.

% is an \emph{annotated source-graph}, or \textbf{\emph{AS-graph}},
% which means it has special node markers called \emph{sources}, drawn
% in red, as well as a \emph{root}, marked with a bold outline. These markers are
% used to combine graphs with the algebra's operations. 
% For instance,
% the \modify{\modsrc} combines the head $\G{sleep}$ with its modifier
% $\G{soundly}$ by plugging its root into the \src{M}-source of
% $\G{soundly}$, see (c). The \app{\src{S}} and \app{\src{O}} operations
% then plug $\G{writer}$ and (c) into the respective sources of
% $\G{want}$. Note that because $\G{want}$ and (c) both have an
% \src{S}-source, \app{\src{O}} merges these nodes, see (d).

\paragraph{Types} The $[\subj]$ annotation at the \obj{}-source of
\G{begin} in Fig.~\ref{fig:as-graphs} is a \textbf{\emph{request}} as to what the \textbf{\emph{type}} of the
\obj{} argument of \G{begin} should be. The type of a graph is the
set of its sources with their request annotations, so the request
$[\subj]$ means that the source set of the argument must be
$\{\subj\}$. Because this is true of \G{glow}, the AM dependency tree is
\emph{well-typed}; otherwise the tree could not be evaluated to a
graph. Thus, the graph constants lexically specify the semantic
valency of each word as well as reentrancies due to e.g.\ control.
%\modify{\modsrc} is allowed if the modifier's type, minus \modsrc, is a subset of the head's type. Here, \modify{\subj} is allowed because the type of \ref{fig:partial-VP}, minus \subj{}, is \emptytype{}, as is the type of \G{fairy}.

If a graph has no sources, we say it has the \emph{empty type}
$\emptytype$; if a source in a graph printed here has no annotation,
it is assumed to have the empty request (i.e.~its argument must have
no sources).

\paragraph{Parsing} \newcite{groschwitz18:_amr_depen_parsin_typed_seman_algeb} use a neural supertagger and dependency parser to predict scores for graph constants and edges respectively. Computing the highest scoring well-typed AM dep-tree is NP-hard; we use their fixed-tree approximate decoder here. %\newcite{lindemann-etal-2020-fast} introduce a transition based model that achieves faster runtimes.
% We build on the
%parser of \newcite{groschwitz18:_amr_depen_parsin_typed_seman_algeb}
% here, since its scoring model is more compatible with our method.

% \todo{talk a sec about parsing and \cite{lindemann-etal-2020-fast} here, and why we use constant/edge-factored model}

% In this work, we develop a simpler yet more complete way of obtaining AM dep-trees, including a heurstics-free way to obtain type requests, and unsupervised learning of source names as latent variables. This saves labour and adds formal guarantees lacking in hand-written heuristics. 

% Groschwitz et al.\ define two approximate
% parsing algorithms, the \sortof{fixed tree decoder}
% that fixes an unlabeled dependency tree first
% %before adding graph constants and operations
% , and a \sortof{projective decoder} that can only derive projective dependency trees and works akin to the CKY algorithm \citep{CockeS70}.

% \paragraph{Notation and terminology.} 
% %\todo{gut hier? Bin nicht 100\% ob wir's brauchen.}\jg{bin nicht sicher wo das hinpasst, kommt Zeit kommt Rat}
% We write $Dom(f)$ for the domain of a partial function $f$, i.e.~the set of objects for which $f$ is defined.

% maybe in the end double check that the below have been introduced right

% $\Omega$ undefined in Sect. 5; set of all source names? or all types?

%\ak{Need to mention fixed-tree decoder (without major explanation), so we can refer to it in Evaluation.}

%\ak{Begriffe ``lexical graph'', ``graph constant'', ``supertag'' klar machen.}

%\ak{ROOT, IGNORE, $\bot$}

%%% Local Variables:
%%% mode: latex
%%% TeX-master: "unsupervised2020"
%%% End:

%% file: constantsFigure.tex
\begin{figure*}[t]
  % the tiny fairy glows
  \centering
\begin{minipage}{0.41\linewidth}
  \begin{subfigure}[b]{0.2\linewidth}
    \centering
    \includegraphics[scale=\defAMRScale]{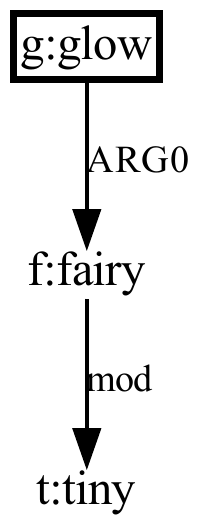}  % exists also triangluar shaped tinyfairyglows.pdf
    \caption{AMR}
    \label{fig:amr-tinyfairyglows}
  \end{subfigure}~
  \begin{subfigure}[b]{0.39\linewidth}
    \centering
    \includegraphics[scale=\defAMRScale]{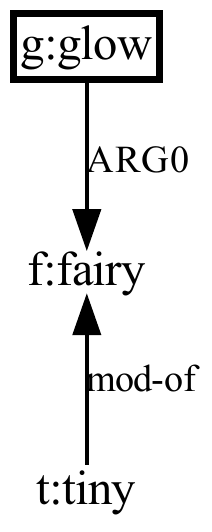}  % exists also triangluar shaped tinyfairyglows.pdf
    \caption{Blob-normalised AMR}
    \label{fig:normalised-tinyfairyglows}
  \end{subfigure}~
  \begin{subfigure}[b]{0.37\linewidth}
    \centering
    \begin{forest}
      for tree={align=center, edge=->, font=\small, l sep=1.2em, l=1.2em, }
      [\depNodeWTypeNode{\H{glow}}{\type{\ps{f}}}{g}
      [\depNodeWTypeNode{\H{fairy}}{\emptytype{}}{f}, name=fairy, edge label = {node [midway, left] {\small\app{\ps{f}}}}
      [\depNodeWTypeNode{\H{tiny}}{\type{\ps{f}}}{t}, edge label = {node [midway, left] {\small\modify{\ps{f}}}}
      ]
      ]
      ]
    \end{forest}
    \caption{Canonical AM tree with types}
    \label{fig:tree-tinyfairyglows}
  \end{subfigure}
  \caption{\sent{The tiny fairy glows}.
%Semantic graph without reentrancies and its canonical AM dep-tree.
}\vspace{-10pt}
\label{fig:noReent}
\end{minipage}\quad~~
% canonical constants with placeholders (C-graph)
\begin{minipage}{0.28\linewidth}
\centering
    \begin{minipage}{\linewidth}
    \centering
    \begin{minipage}{0.4\linewidth}
    \begin{flushright}
      \includegraphics[scale=\defAMRScale]{pics/fairy.pdf}
    \end{flushright}
    \end{minipage}
      \begin{minipage}{0.4\linewidth}
        \small \H{fairy}
      \end{minipage}
    \end{minipage}
    \vspace{0.5em}
    
  \begin{minipage}[b]{0.2\linewidth}
    \includegraphics[scale=\defAMRScale]{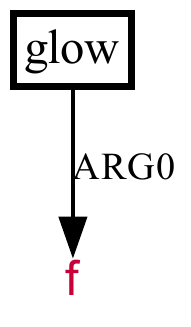}
    \caption*{\small \H{glow}}
  \end{minipage}~
  \begin{minipage}[b]{0.2\linewidth}
    \includegraphics[scale=\defAMRScale]{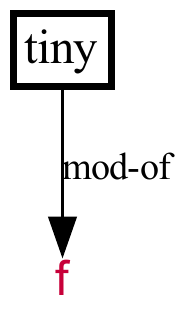}
    \caption*{\small \H{tiny}}
  \end{minipage}~
  \begin{minipage}[b]{0.24\linewidth}
    \includegraphics[scale=\defAMRScale]{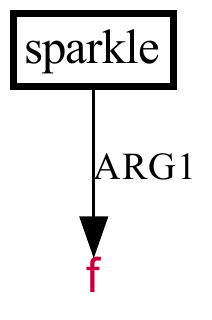}
    \caption*{\small \H{sparkle}}
  \end{minipage}~
  \begin{minipage}[b]{0.24\linewidth}
    \includegraphics[scale=\defAMRScale]{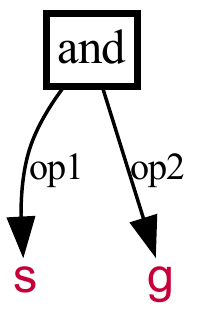}
    \caption*{\small \H{and}}
  \end{minipage}
  \caption{Canonical constants.
  %Note graphs with no sources do not change, so e.g. \G{fairy}=\H{fairy}=\Hhat{fairy}
  }
  \label{fig:placeholder-graphs}
\end{minipage}\quad~
% Non-canonical constants with placeholder sources (H-graph)
\begin{minipage}{0.23\linewidth}
  \centering
  \begin{minipage}[b]{0.45\linewidth}
    \includegraphics[scale=\defAMRScale]{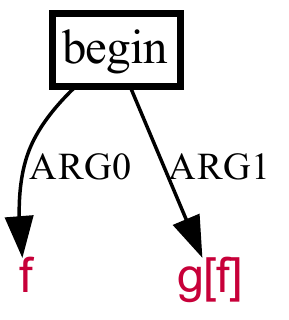}
    \caption*{\small \Hhat{begin}}
  \end{minipage}~
  \begin{minipage}[b]{0.45\linewidth}
    \includegraphics[scale=\defAMRScale]{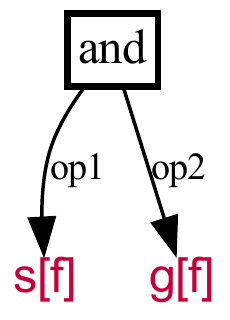}
    \caption*{\small \Hhat{and}}
  \end{minipage}
  \caption{Non-canonical constants with placeholder sources.}
  % \caption{Constants with placeholder sources}
  \label{fig:hbars}
\end{minipage}
\end{figure*}

%% file: decomposition-new.tex
\section{Decomposition algorithm}\label{sec:decomp}

 The central challenge of compositional methods lies in the fact that the compositional structures are not provided in the graphbanks.
Existing AM parsers \cite{groschwitz18:_amr_depen_parsin_typed_seman_algeb, lindemann-etal-2019-compositional, lindemann-etal-2020-fast} use hand-built heuristics to extract AM dep-trees for supervised training from the graphs in the graphbank. These heuristics require extensive expert work, including graphbank-specific decisions for source allocations and graphbank- and phenomenon-specific patterns to extract type requests for reentrancies. 
In this section we present a simpler yet more complete method for obtaining the basic structure of an AM dep-tree for a given semantic graph $G$ (for \emph{decomposing} the graph), with much reduced reliance on heuristics. We will learn meaningful source names jointly with training the parser in \myref{sec:auto} and \myref{sec:neural}.

\textbf{Notation.}
We treat graphs as a quadruple $G = \la N_G, r_G, E_G, L_G \ra$, where the nodes $N_G$ are arbitrary objects (in the examples here we use lowercase letters), $r_G\in N_G$ is the root, $E_G\subseteq N_G\times N_G$ is a set of directed edges, and $L_G$ is the labelling function for the nodes and edges. For example in Fig.~\ref{fig:amr-tinyfairyglows}, the node $g$ is labeled ``glow''. The node identities are not relevant for graph identity or evaluation measures, but allow us to refer to specific nodes during decomposition. We formalize AM dep-trees as similar quadruples. Note that our example graphs are all AMRs, but our algorithms apply unchanged to all graphbanks

\subsection{Basic transformation to AM dep-trees}
\label{sec:basic-transformation}
Let us first consider the case where the semantic graph $G$ has no reentrancies, like in Fig.~\ref{fig:amr-tinyfairyglows}. The first step in obtaining the AM dep-tree for $G$ is to obtain the basic shape of the constants.
We let each graph constant contain exactly one labeled node. Each edge belongs to the constant of exactly one node. The edges in the constant of a node are called its \textbf{\emph{blob}} \citep{groschwitz-etal-2017-constrained}; the blobs partition the edge set of the graph. For example, the blobs of the AMR in Fig.~\ref{fig:amr-tinyfairyglows} are \nn{g} plus the \el{\textsc{arg0}} edge, \nn{t} plus the \el{mod} edge, and \nn{f}.
We normalise edges so that they point away from the node to whose blob they belong, like in Fig.~\ref{fig:normalised-tinyfairyglows}, where the \el{mod} edge is reversed and grouped with the node \nn{t} to match \H{tiny} in Fig.~\ref{fig:placeholder-graphs}. We add an \textit{-of} suffix to the label of reversed edges. From here on, we assume all graph edges to be normalised this way.

Heuristics for this partition of edges into blobs are simple yet effective. Thus, this is the only part of this method where we still rely on graphbank-specific heuristics. (We use the same blob heuristics as \newcite{lindemann-etal-2019-compositional} in our experiments).

% For each edge from a node $n$ to a node $m$ we then have to decide whether the edge goes into the constant of $n$ or $m$. In other words, we decide whether $n$ fills a slot of $m$ or vice versa\todo{use the word blobs}.
Once the decision of which edge goes in which blob is made,
% \todo{should we say here that this is determined heuristically for the whole sembank?}
we obtain \textbf{\emph{canonical constants}}, which are single node constants using \textbf{\emph{placeholder source names}} and the empty request at every source; see e.g.~\H{glow} in Fig.~\ref{fig:placeholder-graphs} (\textit{P} for `placeholder'). 
Placeholder source names are graph-specific source names: for a given argument slot in a constant, let \nn{n} be the node that eventually fills it in $G$; we write \ps{n} for the placeholder source in that slot. For example in the AM dep-tree in Fig.~\ref{fig:tree-tinyfairyglows} the source \ps{f} in \H{glow} (Fig.~\ref{fig:placeholder-graphs}) gets filled by node \nn{f} in the AMR in Fig.~\ref{fig:normalised-tinyfairyglows}. These placeholder sources
%are not suited to generalize across the corpus or carry linguistic meaning, but they
are unique within the graph, allowing us to track source names through the AM dep-tree.
When we restrict ourselves to the canonical constants, in a setting without reentrancies, the compositional structure is fully determined by the structure of the graph:

\begin{lemma}\label{lem:treeAM}
For a graph $G$ without reentrancies, given a partition of $G$ into blobs, there is exactly one AM dep-tree $C_G$ with canonical constants that evaluates to $G$.
\end{lemma}

We call this AM dep-tree the \textbf{\emph{canonical AM tree}} $C_G=\la N_G, r_G, E_C, L_C\ra$ of $G$. Fig.~\ref{fig:tree-tinyfairyglows} shows the canonical AM tree for the graph in Fig.~\ref{fig:normalised-tinyfairyglows}, using the canonical constants in Fig.~\ref{fig:placeholder-graphs}. The canonical AM tree uses the same nodes and root as $G$, and essentially the same edges, but all edges point away from the root, forming a tree. Each node is labeled with its canonical constant. Each edge $\edge{n}{m}{}\in E_C$ is labeled \app{\ps{m}} if the corresponding edge in the graph has the same direction, and is labeled \modify{\ps{n}} if there is instead an edge $\edge{m}{n}{}$ in $G$.

% with reentrancies
\subsection{Reentrancies and types}

\begin{figure}
  \centering
  % AMR
  \vspace{5pt}
  \begin{subfigure}[t]{0.24\linewidth}
    \centering
    %\resizebox{\textwidth}{!}{
    \includegraphics[scale=\defAMRScale]{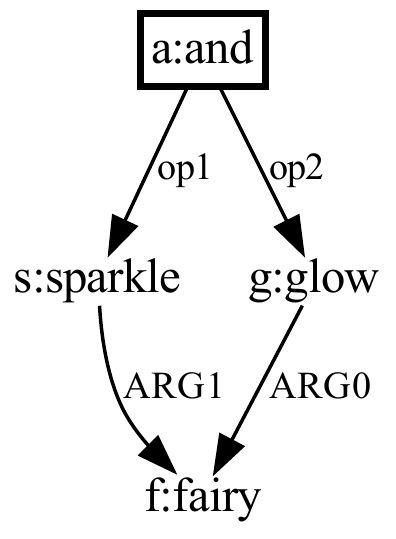}
    %}
    \caption{AMR $G$}
    \label{fig:amr-sparkle-and-glow}
  \end{subfigure}
  % Unrolled AMR
  \begin{subfigure}[t]{0.25\linewidth}
    \centering
    %\resizebox{\textwidth}{!}{
    \includegraphics[scale=\defAMRScale]{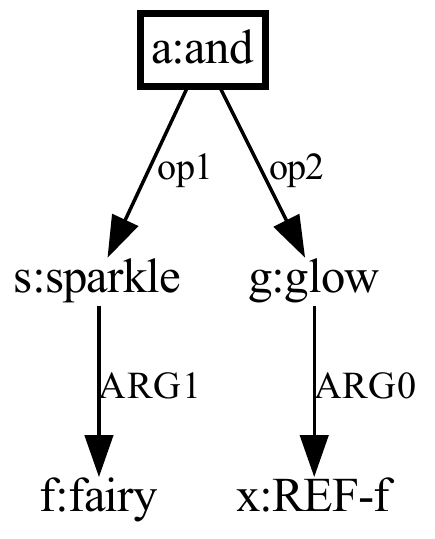}
    %}
    \caption{Unrolled $U$}
    \label{fig:unrolled}
  \end{subfigure}
  % Canonical AM tree
  \begin{subfigure}[t]{0.48\linewidth}
    \resizebox{\textwidth}{!}{
    \centering
    \begin{forest}
    for tree={align=center, edge=->, font=\small, l sep=1.2em, l=1.2em, }
    [\depNodeWTypeNode{\H{and}}{\type{\ps{s,g}}}{a}, name=and        
    [\depNodeWTypeNode{\H{sparkle}}{\type{\ps{f}}}{s}, name=sparkle, edge = blue, edge label = {node [midway, right] {\small \app{\ps{s}}}}
       [\depNodeWTypeNode{\H{fairy}}{\emptytype{}}{f}, name=fairy, edge = blue, edge label = {node [midway, right] {\small \app{\ps{f}}}}
       ]
    ]
    [\depNodeWTypeNode{\H{glow}}{\type{\ps{f}}}{g}, name=glow, edge = dkgreen, edge label = {node [midway, right] {\small \app{\ps{g}}}}
        [\depNodeWTypeNode{\refnode{\nn{f}}}{\emptytype{}}{x}, name=ref, edge = dkgreen,  edge label = {node [midway, right] {\small \app{\ps{f}}}}
        ]
        ]
        ]
    \end{forest}
    }
    \caption{Canonical AM tree $C_U$}
    \label{fig:canonical}
  \end{subfigure}
  
  % Partial result
  \begin{subfigure}[b]{0.26\linewidth}
    \centering
    \resizebox{\textwidth}{!}{
    \includegraphics[scale=\defAMRScale]{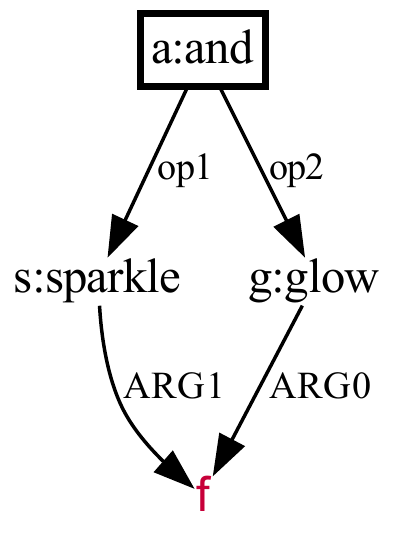}
    }
    \caption{Partial result}
    \label{fig:andPartial}
  \end{subfigure}\quad
  % Resolved tree
  \begin{subfigure}[b]{0.69\linewidth}
    \centering
    \resizebox{\textwidth}{!}{
    \begin{forest}
    for tree={align=center, edge=->, font=\small, l sep=1.2em, l=1.2em, }
    [\depNodeWTypeNode{\Hhat{and}}{\type{\ps{s{\color{purple}[f]},g{\color{purple}[f]}}}}{a}, name=begin
      [\depNodeWTypeNode{\H{sparkle}}{\type{\ps{f}}}{s}, edge label = {node [midway, right] {\small \app{\ps{s}}}}        
      ]
      [\depNodeWTypeNode{\H{glow}}{\type{\ps{f}}}{g}, edge label = {node [midway, right] {\small \app{\ps{g}}}}
      ]
      [\color{purple}\depNodeWTypeNode{\H{fairy}}{\emptytype{}}{f}, name=fairy, edge = purple, edge label = {node [midway, right] {\small \app{\ps{f}}} }
      ]
      ]
    \end{forest}
    }
    % \caption{AM dep-tree with place-holder sources}
    \caption{Resolved AM dep-tree $T$ for (a); changes with respect to (c) in purple}
    \label{fig:am-placeholder-reent}
  \end{subfigure}
  \caption{Analysis for \sent{The fairy sparkles and glows}.}
  \label{fig:percolation}
\end{figure}

Finding AM dep-trees for graphs \emph{with} reentrancies, like in Fig.~\ref{fig:amr-sparkle-and-glow}, is more challenging. To solve the problem in its generality, we first \emph{unroll} the graph as in Fig.~\ref{fig:unrolled}, representing the reentrancy at \nn{f} not directly, but with a \textit{reference node} with label \refnode{\nn{f}}.
%, akin to e.g.~the Penman notation for AMR graphs.
Merging this REF-node with the node \nn{f} it refers to yields the original graph again. (See \S\ref{sec:unrolling} for our unrolling algorithm.) An unrolled graph $U$ shares its non-REF-nodes with the original graph $G$. REF-nodes are always leaves.
%We call $\{n\in N_G ~|~ \exists m\in N_U$ s.t. $L_U(m) = \refnode{n}\}$ the set of all \textbf{\textit{copied nodes}} in $U$. TODO

% the fairy sparkles and glows

\input{algorithmResolving}

We then obtain a canonical AM-tree $C_U$ for the unrolled graph $U$ as in \S\ref{sec:basic-transformation} (see Fig.~\ref{fig:canonical}), but \refnode{n} nodes fill \ps{n}-sources; 
e.g.~\nn{x} has an incoming \app{f} edge here.
%e.g.\ \H{glow} has type \type{\ps{f}}, not \type{\ps{x}}.
$C_U$ evaluates to $U$, not to $G$; we obtain an AM dep-tree that evaluates to  $G$ through a process called \emph{resolving} the reentrancies, which removes all REF-nodes and instead expresses the reentrancies with the AM type system.

%, at this stage in $C_U$ notated with REF-nodes, into the types of the graph constants labelling the nodes of the tree.

Fig.~\ref{fig:am-placeholder-reent} shows the result $T$ of applying this resolution process to $C_U$ in Fig.~\ref{fig:canonical}.
%\todo{Canonical constants with no non-empty requests are unchanged by Alg.~\ref{alg:percol}, so e.g. \Hhat{glow}=\H{glow} and \Hhat{fairy}=\H{fairy}.}
In $T$, the \ps{s} and \ps{g} sources of the graph \Hhat{and} (see Fig.~\ref{fig:hbars}) each have a request $[\ps{f}]$ that signals that the \ps{f} sources of \H{sparkle} and \H{glow} are still open when these graphs combine with \Hhat{and}, yielding the partial result in Fig.~\ref{fig:andPartial}. Since identical sources merge in the AM algebra, Fig.~\ref{fig:andPartial} has a  single \ps{f}-source slot. Into this slot, \H{fairy} is inserted to yield the original graph $G$ in Fig.~\ref{fig:amr-sparkle-and-glow}, and we have obtained the reentrancy without using a REF-node. \nn{f} is now a child of \nn{a} in $T$; we call \nn{a} the \textbf{\emph{resolution target}}  of \nn{f}, \resolutionTarget{f}. In general the resolution target of a node \nn{n} is the lowest common ancestor of \nn{n} and all nodes labeled \refnode{n}.

Thus, to resolve the graph, we (a) add the necessary type requests to account for sources remaining open until they are merged at the resolution target and (b) make each node a dependent of its resolution target and remove all \refnode{}nodes. Algorithm~\ref{alg:percol} describes this procedure. It uses the idea of an \nn{n}-\textbf{\emph{resolution path}}, which is a path between a node \nn{n} or a \refnode{n} node and its resolution target. In Fig.~\ref{fig:canonical}, there are two \nn{f}-resolution paths: one in blue between \nn{f} and its resolution target \nn{a}, and one in green between the \refnode{f} node \nn{x} and its resolution target \nn{a}. Further, \lexicaltype{\nn{n}} is the type of the graph constant in $T$ for a node \nn{n}  and \termtype{\nn{n}} is the type of the result of evaluating the subtree below \nn{n} in $T$.

In the example, Algorithm~\ref{alg:percol} iterates over all edges in both resolution paths (Line~\ref{algline:percol:edgeloop}; the order of these iterations does not impact the result). For the two bottom edges \edge{s}{f}{\app{\ps{f}}} and \edge{g}{x}{\app{\ps{f}}}, Line~\ref{algline:percol:bottomedge} applies. Since the subtree rooted at \nn{f} evaluates to a constant with empty type, no actual changes are made here (\termtype{y} can be non-trivial from resolution paths handled previously). For the two upper edges \edge{a}{s}{\app{\ps{s}}} and \edge{a}{g}{\app{\ps{g}}}, Line~\ref{algline:percol:middleedge} applies, adding \ps{f} to the requests at \ps{s} and \ps{g} in the constant at \nn{a}.
%This corresponds to the \ps{f} sources of \Hhat{sparkle} and \Hhat{glow} remaining open when they are combined with \Hhat{and}.
In Line~\ref{algline:percol:moveEdge}, \nn{f} gets moved up to become a child of its resolution target \nn{a} and in Line~\ref{algline:percol:delREF} the \refnode{f} node \nn{x} gets removed, yielding $T$ in Fig.~\ref{fig:am-placeholder-reent}.
Algorithm~\ref{alg:percol} is correct in the following precise sense:

\begin{theorem} \label{thm:mods}
Let $G$ be a graph, let $U$ be an unrolling of $G$, let $C_U$ be the canonical AM-tree of $U$, and let $T$ be the result of applying Algorithm \ref{alg:percol} to $C_U$. Then $T$ is a well-typed AM dep-tree that evaluates to $G$ iff for all $\nn{y}\in N_G$, for all \nn{y}-resolution paths $p$ in $C$,
\begin{enumerate}
    \vspace{-5pt}\item the bottom-most edge \edge{n}{m}{} of $p$ (i.e.~\nn{m} is \nn{y} or labeled \refnode{y}) does not have a \modify{} label, and
    %If $n\in N_C$ is a copied node, no $n$-node has a \modify{} parent in $C$, and \todo{case $n=\text{LCA}\of{n}$; refer to bottom-most edge of $y$-resulution path}
    \vspace{-5pt}\item for all $y$-resolution paths $p$ in $C$, if %$\exists s$ s.t. 
    \edge{n}{m}{\modify{}} $\in p$, $n,m\neq y$, then there is a directed path in $G$ from $n$ to $y$.
\end{enumerate}
%If there exists a well-typed AM dep-tree that evaluates to $G$, then there exists an unrolling $U$ such that (1) and (2) are satisfied.
\end{theorem}

Condition (1) captures the fact that moving \modify{} edges in the graph changes the evaluation result (the modifier would attach at a different node) and Condition (2) the fact that modifiers are not allowed to add sources to the type of the head they modify. 

Algorithm~1 does not yield \emph{all} possible AM dep-trees;
%for example, the resolution target could be higher up in the tree.
in Appendix~B, %\ref{app:percolations}, 
we present an algorithm that yields all possible AM dep-trees (with placeholder sources) for a graph. However, we find in practice that Algorithm~1 almost always finds the best linguistic analysis; i.e.~reasons to deviate from Algorithm~1 are rare (we estimate that this affects about $1\%$ of nodes and edges in the AM dep-tree). We leave handling these rare cases to future work.

\subsection{Unrolling the graph} \label{sec:unrolling}

\input{algorithmUnroll}

To obtain an unrolled graph $U$,
%that satisfies the conditions of Theorem~\ref{thm:mods}
we use Algorithm~\ref{alg:unroll}. The idea is to simply expand $G$ through breadth-first search, creating REF-nodes when we encounter a node a second time. We use separate queues $F$ and $B$ for forward and backward traversal of edges, allowing us to avoid traversing edges backwards wherever possible, since that would yield \modify{} edges in the canonical AM-tree $C_U$, which can be problematic for the conditions of Theorem~\ref{thm:mods}. And indeed, we can show that whenever there is an unrolled graph $U$ satisfying the conditions of Theorem~\ref{thm:mods}, Algorithm~\ref{alg:unroll} returns one.

Algorithm~2 does not specify the order in which the incident edges of each node $n$ are added to the queues, leaving an element of choice. However, we find that nearly all of these choices are unified later in the resolution process; meaningful choices are rare. For example in Fig.~\ref{fig:unrolled}, \nn{f} and \nn{x} may be switched, but Algorithm~\ref{alg:percol} always yields the AM dep-tree in Fig.~\ref{fig:am-placeholder-reent}. In practice, we execute Algorithm~2 with arbitrary queueing order, and follow it with Algorithm~1. The AM dep-tree we obtain is guaranteed to be a decomposition of the original graph whenever one exists:

% Algorithm~2 does not specify the order in which the incident edges of each node $n$ are added to the queues, leaving an element of choice. However, we find that nearly all of these choices are unified later in the resolution process; meaningful choices are rare. For example in Fig.~\ref{fig:unrolled}, \nn{f} and \nn{x} may be switched, but Algorithm~\ref{alg:percol} always yields the AM dep-tree in Fig.~\ref{fig:am-placeholder-reent}. \todo{Further, changes like the ones made in Algorithm~1 can apply beyond the necessity of resolving reentrancies; for example, the resolution target could be any common ancestor, not just the lowest one.} In Appendix~\ref{app:percolations}\todo{}, we present an algorithm that yields \emph{all} possible AM dep-trees (with placeholder sources) for a graph. Here, we obtain one fixed AM dep-tree by executing Algorithm~2 with arbitrary queueing order, followed by Algorithm~1. We find that cases where the result does not coincide with our manual analysis is rare, affecting about \todo{}\% of graph constants and operations. We leave treating those rare cases to future work. The tree we obtain is a decomposition of the original graph whenever one exists:

\begin{theorem}\label{thm:correct}
Let $G$ be a graph partitioned into blobs. If there is a well-typed AM dep-tree $T$, using that blob partition, that evaluates to $G$, then Algorithm~2 (with any queueing order) and Algorithm~1 yield such a tree.
\end{theorem}

%% file: algorithmResolving.tex
% more Pythony
\SetStartEndCondition{ }{}{}%
\SetKwProg{Fn}{def}{\string:}{}\SetKwFunction{Range}{range}%
\SetKw{KwTo}{in}\SetKwFor{For}{for}{\string:}{}%
\SetKwIF{If}{ElseIf}{Else}{if}{:}{elif}{else:}{}%
\SetKwFor{While}{while}{:}{fintq}%

\begin{algorithm}[t]
%\KwResult{an AM dep-tree or some unresolvable mess}
 $T \gets{}$ the canonical AM-tree $C_U$ of an unrolling $U$ of $G$\;
 $R\gets{\{n\in N_G ~|~ \exists\ \text{\refnode{n} node in $U$}\}}$\;
 \While{$R\neq \emptyset$}{
  Pick a $y\in R$ s.t. there is no $x\in R$, $x\neq y$, with $y$ on an $x$-resolution path\;
  \For{$p \in y$-resolution paths}{
  \For{ $\edge{n}{m}{\app{}} \in p$\label{algline:percol:edgeloop}}{
  \eIf{\text{\nn{m} is \nn{y} or labeled \refnode{y}}}{
   Add $\termtype{y}$ to the request at \ps{y} in \lexicaltype{$n$}\;\label{algline:percol:bottomedge}
   }{
   Add \ps{y}[$\termtype{y}$] to the request at \ps{m} in \lexicaltype{$n$}\label{algline:percol:middleedge}\;
  }
  }}
  Move the subtree of $T$ rooted at $y$ up to be an \app{\ps{y}} daughter of \resolutionTarget{y}, unless $\resolutionTarget{y}=y$\label{algline:percol:moveEdge}\;
  Delete all \refnode{y} nodes from $T$\label{algline:percol:delREF}\;
  $R \gets R -\{y\}$
 }
 \Return{$T$}
 \caption{Reentrancy resolution}\vspace{-5pt}
 \label{alg:percol}
\end{algorithm}

%% file: algorithmUnroll.tex
\begin{algorithm}[t]
\KwIn{Graph $G$}
 $F,B \gets{}$ empty FIFO queues\;
 $U\gets{}$ empty graph\;
 add $\rt_G$ to $U$, add outgoing edges of $\rt_G$ to $F$ and incoming edges of $\rt_G$ to $B$\;
 \While{$F\cup B\neq \emptyset$}{
  \eIf(\tcp*[f]{traverse forward}){$F\neq\nul$}{
   $e \gets{}$ $F$.pop\;
   $n \gets{}$ $e$.target\;
   }(\tcp*[f]{traverse backward}){
   $e \gets{}$ $B$.pop\;
   $n \gets$ $e$.origin\;
  }
  Mark $e$ as traversed\;
  \eIf{$n\not\in N_U$}
  {add $n,e$ to $U$\;
  add untraversed outgoing edges of $n$ to $F$ and incoming to $B$
  }
  {
  add new $x$ to $N_U$; $L(x)=\refnode{n}$\;
  add $e'$ to $E_U$ where $e'$ is just like $e$ except with $x$ in place of $n$}
  }
 \Return{$U$}
 \caption{Unrolling}
 \label{alg:unroll}
\end{algorithm}

%% file: automata.tex
\section{Tree automata for source names}\label{sec:auto}

We have now seen how, for any graph $G$, we obtain a unique AM dependency tree $T$.
This tree represents the compositional structure of $G$, but it still contains placeholder source names.
We will now show how to automatically choose source names. These names should be consistent across the trees for different sentences; this yields reusable graph constants, which capture linguistic generalizations and permit more accurate parsing.
But the source names must also remain consistent \emph{within} each tree to ensure that the tree still evaluates correctly to $G$; for instance, if we replace the placeholder source \ps{f} in \H{glow} in Fig.~\ref{fig:am-placeholder-reent} by \obj, but we replace \ps{f} in \Hhat{and} by \subj, then the AM dep-tree would not be well-typed because the request is not satisfied. 

We
therefore
proceed in two steps. In this section, we represent all internally consistent source assignments compactly with a tree automaton. In  \S \ref{sec:neural}, we then
learn to select globally reusable source names jointly with training the neural parser.

\paragraph{Tree automata.} A (bottom-up) tree automaton \cite{ComonDGJLTL07} is a device for compactly describing a language (set) of trees.
It processes a tree bottom-up, starting at the leaves, and nondeterministically assigns states from a finite set to the nodes.
A rule in a tree automaton has the general shape
$\arule{q}{f}{q_1,\dots,q_n}$.
If the automaton can assign the states $q_1,\ldots,q_n$ to the children of a node $\pi$ with node label $f$, this rule allows it to assign the state $q$ to $\pi$. The automaton \emph{accepts} a tree if it can assign a \emph{final state} to the root node. Tree automata can be seens as generalisation of parse charts.

\paragraph{General construction.} Given an AM dependency tree $T$ with placeholders, we construct a tree automaton that accepts all well-typed variants of $T$ with consistent source assignments. More specifically, let \sources{} be a finite set of reusable source names; we will use $\sources{}=\left\{\subj,\obj,\modsrc\right\}$ here, evoking subject, object, and modifier. The automaton will keep track of \emph{source name assignments}, i.e.\ of partial functions $\phi$ from placeholder source names into \sources{}. Its rules will ensure that the functions $\phi$ assign source names consistently.

\begin{figure}
    \centering
    % AM tree with placeholder sources
    \begin{subfigure}[b]{0.2\linewidth}
    \hspace{-37pt}
    \begin{minipage}{\linewidth}
    \resizebox{70pt}{!}{
    \begin{forest}
    for tree={align=center, edge=->, font=\small, l sep=1.8em, l=1.8em, }
    [\quad\quad\depNodeWType{\H{fairy}}{\emptytype{}}
        [\quad\quad\depNodeWType{{}\Hhat{begin}}{\type{\ps{f, g\type{\ps{f}}}}}, edge label = {node [midway, right] {\small\modify{\ps{f}}}}
            [\quad\quad\depNodeWType{\H{glow}}{\type{\ps{f}}}, edge label = {node [midway, right] {\small\app{\ps{g}}}}
            ]
        ]
    ]
    \end{forest}}
    \end{minipage}
    \centering
    \caption{}
    \label{fig:amr-relative-tree1}
    \end{subfigure}
    % binarization
    \begin{subfigure}[b]{0.34\linewidth}
    \centering
    \resizebox{\textwidth}{!}{
    \begin{forest}
    for tree={s sep=0.0em, inner sep=0.2em, l sep=0.6em
    , l=0.6em}
    [\modify{\ps{f}}
        [\small \H{fairy}]
        [\app{\ps{g}}, l=0.5em
            [\small \Hhat{begin}] % \scriptsize
            [\small \H{glow}]
        ]
    ]
    \end{forest}
    }
    \caption{}
    \label{fig:binary}
    \end{subfigure}
    \hspace{-15pt}
    % Automaton run
    \begin{subfigure}[b]{0.49\linewidth}
    \centering
    % \small
    \resizebox{\textwidth}{!}{
    \begin{forest}
    for tree={s sep=0.0em, inner sep=0.2em, l sep=0.6em
    , l=0.6em, align=left}
    [{\modify{\subj}\\
            \color{dkgreen}$\astate{\ensuremath{\epsilon},\{\}}$
            }
        [{\small \G{fairy}\\
            \color{dkgreen}$\astate{0,\left\{ \right\}}$
            }]
        [{\quad\app{\obj}\\
                \color{dkgreen}$\astate{1,\left\{\substack{\ps{g}\mapsto\obj\\\ps{f}\mapsto\subj} \right\}}$
                }
            [{\small\quad\quad \G{begin}\\
            \color{dkgreen}$\astate{10,\left\{\substack{\ps{g}\mapsto\obj\\\ps{f}\mapsto\subj} \right\}}$
            }] % \scriptsize
            [{\small \G{glow}\\
            \color{dkgreen}$\astate{11,\left\{\substack{\ps{f}\mapsto\subj} \right\}}$
            }]
        ]
    ]
    \end{forest}
    }
    \caption{}
    \label{fig:automaton}
    \end{subfigure}
    
    \caption{(a) AM dep-tree with placeholder sources for the graph in Fig.~\ref{fig:amr-fairy}, (b) its binarization $B$ and (c) example automaton run (states in green).}
    \label{fig:binaryTree}
\end{figure}
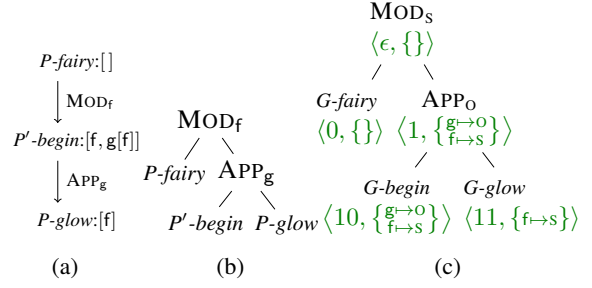

We start by binarizing $T$ into a binary tree $B$, whose leaves are the graph constants in $T$ and whose internal nodes correspond to the edges of $T$; the binarized tree for the dependency tree in Fig.~\ref{fig:amr-relative-tree1} is shown in Fig.~\ref{fig:binary}. We then construct a tree automaton $A_B$ that accepts
binarized trees which are isomorphic to $B$, but whose node labels have been replaced by graph constants and operations with reusable source names. The states of $A_B$ are of the form \astate{\pi,\phi}, where $\phi$ is a source name assignment and $\pi$ is the address of a node in $B$. Node addresses $\pi\in\N^*$ are defined recursively: the root has the empty address $\epsilon$, and the $i$-th child of a node at address $\pi$ has address $\pi i$. The final states are all states with $\pi=\epsilon$, indicating that we have reached the root.

\paragraph{Rules.} The automaton $A_B$ has two kinds of rules. \emph{Leaf rules} choose injective source name assignments for constants; there is one rule for every possible assignment at each constant.
That is, for every graph constant $H$ at an address $\pi$ in $B$, the automaton $A_B$ contains all rules of the form
$$G\mapsto \astate{\pi,\phi}$$
where $\phi$ is an injective map from the placeholder sources in $H$ to \sources, and $G$ is the graph constant identical to $H$ except that each placeholder source $s$ in $H$ has been replaced by $\phi(s)$.

For example, the automaton for Fig.~\ref{fig:binary} contains the following rule:
$$\G{begin} \rightarrow \astate{00,\left\{\ps{g}\mapsto\obj,\ps{f}\mapsto\subj\right\}}$$
Note that this rule uses the node label $\G{begin}$ with the reusable source names, not the graph constant \Hhat{begin} in $B$ with the placeholders.

In addition, \emph{operation rules} percolate source assignments from children to parents. Let $\app{x}$ for some placeholder source $x$ be the operation at address $\pi$ in $B$. Then $A_B$ contains all rules of the form
$$\arule{\astate{\pi,\phi_1}}{\app{\phi_1\of{x}}}{\astate{\pi0,\phi_1}, \astate{\pi1, \phi_2}}$$
as long as $\phi_1$ and $\phi_2$ are identical where their domains overlap, i.e.\ they assign consistent source names to the placeholders. The rule passes $\phi_1$ on to its parent. The assignments in $\phi_2$ are either redundant, because of overlap with $\phi_1$, or they are no longer relevant because they were filled by operations further below in the tree.
%because of details of the AM type system \cite{groschwitz-etal-2017-constrained}. 
The \modify{} case works out similarly.
% \modify{} works out similarly.

In the example, $A_B$ contains the rule
$$\app{\obj}\of{\astate{10,\phi_b}, \astate{11,\phi_g}}
\rightarrow \astate{1,\phi_b}$$
where $\phi_b=\left\{\ps{g}\mapsto\obj,\ps{f}\mapsto\subj\right\}$ and $\phi_g=\left\{\ps{f}\mapsto\subj\right\}$,
because $\phi_b$ and $\phi_g$ agree on $\ps{f}$. A complete accepting run of the automaton is shown in Fig.~\ref{fig:automaton}.

The automaton $A_B$ thus constructed accepts the binarizations of all well-typed AM dependency trees with sources in \sources{} that match $T$.

%% file: neural.tex
\section{Joint learning of compositional structure and parser}\label{sec:neural}

As a final step, we train the neural parser of  \newcite{groschwitz18:_amr_depen_parsin_typed_seman_algeb} directly on
the tree automata. For each position $i$ in the sentence, the parser predicts a score $\graphCost{i}{G}$ for each graph constant $G$,
and for each pair $i,j$ of positions and operation $\oper$, it predicts an edge score $\edgeCost{i}{j}{\oper}$.

%By design,
The tree automata are factored the same way, in that they have one rule per graph constant and per dependency edge. As a result, we get a one-to-one correspondence between parser scores and automaton rules when aligning automata rules to words via the words' alignments to graph nodes.

We thus take the neural parser scores as \emph{rule weights} $\score\of{r}$ for rules $r$ in the automaton.
% The tree automata are constructed such that each rule makes a decision on either one graph constant or one dependency edge, and we align rules to words via words' alignments to graph nodes.
% We can thus take the neural scores of these constants and edges as \emph{rule weights} $\score\of{r}$ for rules $r$ in the automaton.
In a weighted tree automaton,
the weight of a tree is defined as the product of the weights of all rules that built it. The \emph{inside score} $I$ of the tree automaton is the sum of the weights of all the trees it accepts.
Computing this sum naively would be intractable, but the inside score can be computed efficiently with dynamic programming.
Our training objective is to maximize the sum of the log inside scores of all automata in the corpus.

The arithmetic structure of computing the inside scores is complex and varies from automaton to automaton, which would make batching difficult. We solve this with the chain rule as follows:
\begin{align*}
    \nabla_\theta\log I=\frac{1}{I}\nabla I 
    &=\frac{1}{I}\sum_{r\in A}\frac{\partial}{\partial \score\of{r}}I\:\:\nabla_\theta\score\of{r}\\
    &=\frac{1}{I}\sum_{r\in A}\alpha\of{r}\nabla_\theta\score\of{r},
\end{align*}
where $\theta$ are the parameters of the neural parser, which determine $c(r)$, and $\alpha\of{r}$ is the \emph{outer weight} of the rule $r$ \cite{eisner2016inside}, i.e.\ the total weight of trees that use $r$ divided by $c(r)$. The outer weight can be effectively computed with the inside-outside algorithm \cite{baker-io}. This occurs outside of the gradient, so we do not need to backpropagate into it. Since the scores $\score\of{r}$ are direct outputs of the neural parser, their gradients can be batched straightforwardly.

%% file: results.tex
\begin{table}[t]
\centering
\small
\begin{tabular}{lrrrr}
Method         & DM            & PAS           & PSD           & AMR            \\ 
\hline
random trees   & 81.1          & 79.0          & 67.8          & 70.8           \\
random weights & 93.0          & 94.4          & 80.0          & 75.0           \\
EM weights     & 93.8          & 94.3          & 81.7          & 75.2           \\
joint neural model (\myref{sec:neural})   & \textbf{94.5} & \textbf{94.8} & \textbf{82.7} & \textbf{76.5} 
\end{tabular}
\vspace{-5pt}
\caption{Baseline comparisons on the development sets (3 source names in all experiments).}\label{tab:methods}
\end{table}

\input{result-table}

\section{Evaluation}\label{sec:results}
\subsection{Setup}

We evaluate parsing accuracy on the graphbanks DM, PAS, and PSD from the
SemEval 2015 shared task on Semantic Dependency Parsing (SDP,
\citet{OepenKMZCFHU15}) and  on the
AMRBank LDC2017T10 \citep{amBanarescuBCGGHKKPS13}. We follow \newcite{lindemann-etal-2019-compositional} in the choice of neural architecture, in particular using BERT \citep{BERT} embeddings, and in the choice of decoder, hyperparameters and pre- and postprocessing (we train the model of \myref{sec:neural} for 100 instead of 40 epochs, since it is slower to converge than supervised training). 
When a graph $G$ is \emph{non-decomposable} using our blob partition, i.e.~if there is no well-typed AM dep-tree $T$ that evaluates to $G$, and so the condition of Theorem~\ref{thm:correct} does not hold, then we remove that graph from the training set. (This does not affect coverage at evaluation time.) This occurs rarely, affecting e.g.~about $1.6\%$ of graphs in the PSD training set.

Like \cite{lindemann-etal-2019-compositional}, we use the heuristic AMR alignments of \cite{groschwitz18:_amr_depen_parsin_typed_seman_algeb}. These alignments can yield multi-node constants. In those cases, we first run the algorithm of Section~\ref{sec:decomp} to obtain an AM tree with placeholder source names, and then consolidate those constants that are aligned to the same word into one constant, effectively collapsing segments of the AM tree into a single constant. We then construct the tree automata of Section~\ref{sec:auto} as normal.

% \paragraph{Baselines.}
\subsection{Results}
We consider three baselines. Each of these chooses a single tree for each training instance from the tree automata and performs supervised training. The \textbf{random trees} baseline samples a tree for each sentence from its automaton, uniformly at random. In the \textbf{random weights} baseline, we fix a random weight for each graph constant and edge label, globally across the corpus, and select the highest-scoring tree for each sentence.  The \textbf{EM weights} baseline instead optimizes these global weights with the inside-outside algorithm.

Table~\ref{tab:methods} compares the baselines and the joint neural method. Random trees perform worst -- consistency across the corpus matters. The difference between random weights and EM is suprisingly small, despite the EM algorithm converging well. The joint neural learning outperforms the baselines on all graphbanks; we analyze this in \S~\ref{sec:linguistic-analysis}. We also experimented with different numbers of sources, finding $3$ to work best for DM, PAS and AMR, and $4$ for PSD (all results in Appendix~%\ref{app:sources}
C).

% moved to appendix
% \paragraph{Number of source names.}
%\todo{shorten or remove?} We also experimented with different numbers of source names in the joint neural method (Table~\ref{tab:sources}). Mostly, three source names were most effective, except for PSD, where four were most effective. Two source names are not enough to model many common phenomena (e.g.~ditransitive verbs, coordination of verbs); graphs containing these phenomena cannot be decomposed with two sources and are removed from the training set, reducing parsing accuracy.
% The higher performance of PSD with four sources may stem from PSD using flat coordination structures which require more source names; although this is also true for AMR where four source names are not beneficial.
%The drop with six source names may come from the fact that the latent space grows rapidly with more sources, making it harder to learn consistent source assignments.

%when more sources are added, the latent space becomes larger and the model may have a harder time finding consistent explanations.
%\todo{quantify decomposability?}

% \paragraph{Comparison to other work.} 
Table~\ref{tab:results} compares the accuracy of our joint model to \newcite{lindemann-etal-2019-compositional} and to the state of the art on the respective graphbanks. Our model is competitive with the state of the art on most graphbanks. In particular, our parsing accuracy is on par with \newcite{lindemann-etal-2019-compositional}, who perform supervised training with hand-crafted heuristics. This indicates that our model learns appropriate source names.

%(L'19) we see that switching from handwritten heuristics to unsupervised learning does not hinder parsing performance on DM and PAS. On PSD and AMR, the %performance is still comparable. This indicates that our method indeed learns appropriate source names.
%; a conclusion that is supported by our manual inspection in \myref{sec:linguistic-analysis}.

%\todo{cut? -- would definitely cut (AK)} The pre-and postprocessing steps of \cite{lindemann-etal-2019-compositional} we use for AMR still rely on their reentrancy heuristics to remove some edges related to coreference (a non-compositional source of reentrancy). We include in brackets the results without this preprocessing step. The drop in performance indicates a potential benefit of extending AM dependency parsing to handle coreference, a process already started by \newcite{anikina-etal-2020-predicting}.

\paragraph{Grahbank-specific pre- and processing.} The pre- and postprocessing steps of \cite{lindemann-etal-2019-compositional} we use still rely on two  graphbank-specific heuristics, that directly relate to AM depenency trees: in PSD, it includes a simple but effective step to make coordination structures more compatible with the specific flavor of application and modification of AM dependency trees. In AMR it includes a step to remove some edges related to coreference (a non-compositional source of reentrancy).

We include in brackets the results \emph{without} those two preprocessing steps. The drop in performance for PSD indicates that while for the most part our method is graphbank-independent, not all shapes of graphs are equally suited for AM dependency-parsing and some preprocessing to bring the graph \sortof{into shape} can still be important. For AMR, keeping the co-reference based edges leads to AM trees that resolve those reentrancies with the AM type system. That is, the algorithm \sortof{invents} ad-hoc compositional explanations for a non-compositional phenomenon, yielding graph constants with type annotations that do not generalize well.
% Keeping the co-reference based edges also leads to more graphs not being decomposeable, forcing us to remove them from the training set.
% \todo{talk about decomposability?}
The corresponding drop in performance indicates that extending AM dependency parsing to handle coreference will be an important future step when parsing AMR; some work in that direction has already been undertaken \citep{anikina-etal-2020-predicting}.

%%% Local Variables:
%%% mode: latex
%%% TeX-master: "unsupervised2020"
%%% End:

%% file: result-table.tex
\begin{table*}[t]
\centering
\resizebox{\textwidth}{!}{
\small
	\begin{tabular}{l|ll|ll|ll|l}
	\toprule
		& \multicolumn{2}{c}{\textbf{DM}} & \multicolumn{2}{c}{\textbf{PAS}} & \multicolumn{2}{c}{ \textbf{PSD} } & \textbf{AMR 17} \\
		& id F & ood F & id F & ood F & id F & ood F & Smatch F  \\
		\midrule
		%\citet{peng17:_deep_multit_learn_seman_depen_parsin} Basic & 89.4 & 84.5 & 92.2 & 88.3 & 77.6 & 75.3 & - & - & - & - \\
		 %\citet{peng17:_deep_multit_learn_seman_depen_parsin} Freda3 & 90.4 & 85.3 & 92.7 & 89.0 & 78.5 & 76.4 & - & - & - & - \\ %[1.0ex]
		%\citet{wang-etal-2019-second} & 94.0 & 89.7 & 94.1 & 91.3 & 81.4 & 79.6 & - & - & - & - \\
		\citet{bertbaseline} & 94.6 & 90.8 & 96.1 & 94.4 & 86.8 & 79.5 & - \\ 
		FG'20 & 94.4 & 91.0 & 95.1 & 93.4 & 82.6 & 82.0 & -  \\
        % \citet{cai20:_amr_parsin} & - & - & - & - & - & - & 80.2  \\
        %\citet{zhang-etal-2019-broad} & 92.2 & 87.1 & - & - & - & - & 77.0\cin{0.1} \\
        \citet{bevilacqua2021one} & - & - & - & - & - & - & 84.5  \\
		
          \midrule[0.11em]
          
		\lastACL, w/o MTL  & 93.9\cin{0.1} & 90.3\cin{0.1} &94.5\cin{0.1} & 92.5\cin{0.1} & 82.0\cin{0.1} &81.5\cin{0.3}  & 76.3\cin{0.2}\\
		%ALT, mit BUG:
		%74.3\cin{0.2} & 75.3\cin{0.2} \\

		%\lastACL, MTL  & 91.2\cin{ 0.1 } & 85.7\cin{ 0.0 } & 92.2\cin{ 0.2 } & 88.0\cin{ 0.3 } & 78.9\cin{ 0.3 } & 76.2\cin{ 0.4 } & 88.2\cin{ 0.1 } & 83.3\cin{ 0.1 } &  (70.4)\footnotemark\\cin{ 0.2} & 71.2\cin{ 0.2} \\
		
        %\lastACL, MTL & 94.1\cin{0.1} & 90.5\cin{0.1} & 94.7\cin{0.1} & 92.8\cin{0.1} & 82.1\cin{0.2} & \textbf{81.6}\cin{0.1} & 90.4\cin{0.1} & 85.2\cin{0.1} & (74.5)$^3$\cin{0.1} & 75.3\cin{0.1} \\
        %\midrule
        %\lastACL\ + CharCNN & 90.5\cin{0.1} & 84.5\cin{0.1} & 91.5\cin{0.1}&		86.5\cin{0.1}	 &	78.4\cin{0.2}	&	74.8\cin{0.2}& 87.7\cin{0.1} & 82.8\cin{0.1} & 70.2\cin{0.4} & 71.4\cin{0.2} \\
        
        %ML: rausgenommen:
        %\midrule
       % \lastACL  + CharCNN & 93.8\cin{0.1}&90.2\cin{0.1}&94.6\cin{0.1} &		92.5\cin{0.1}	&	81.9\cin{0.1}	&	\textbf{81.5}\cin{0.2}&90.2\cin{0.1}& 85.0\cin{0.1}&75.1\cin{0.2}&76.1\cin{0.1} \\
        
          %Projective parser & \\
          
          \midrule[0.11em]
        
          This work & 94.2\cin{0.0} & 90.2\cin{0.1} & 94.6\cin{0.0} &
                                                                      92.7\cin{0.1} & 81.4\cin{0.1} (75.8\cin{0.1}) & 80.7\cin{0.4} (74.1\cin{0.1}) & 75.1\cin{0.2} (74.2\cin{0.3})\\[0.9ex]

                                                                      % PSDid \todo{73}
                                                                      % PSD ood  \todo{72}
          % AMR17 (74.2\cin{0.3})
          
          %\midrule
         
        \bottomrule
	\end{tabular}}
      %\strut\\[-6ex]
\caption{Semantic parsing accuracies (id = in domain test set; ood = out of domain test set). Results for our work are averages of three runs with standard deviations. \lastACL\ are results of \citet{lindemann-etal-2019-compositional} with fixed tree decoder (incl.~post-processing bugfix for AMR as per \citet{lindemann-etal-2020-fast}). FG'20 is \citet{lys-2020-transition}.}
\vspace{-10pt}
\label{tab:results}
\end{table*}

%%% Local Variables:
%%% mode: latex
%%% TeX-master: "am-parsing-20"
%%% End:

%% file: linguistic-analysis.tex
\section{Linguistic Analysis}
\label{sec:linguistic-analysis}

As AM parsing is inherently interpretable, we can explore linguistic properties of the learned graph constants and trees. We find that the neural method makes use of both syntax and semantics.

%Our neural method replaces the linguist by learning its own graph constants and AM trees, but does what it learns make linguistic sense? We use the inherent interpretability of AM parsing to explore this question, finding that the neural method makes use of syntax in addition to semantics.

We compute for each sentence in the training set the best tree from its tree automaton, according to the neural weights of the best performing epoch. We then sample trees from this set for hand-analysis (see Appendix A), to examine whether the model learned consistent sources for subjects and objects. We find that while the EM method uses highly consistent graph constants and AM operations, the neural method, which has access to the strings, sacrifices some graph constant and operation consistency in favour of \textit{syntactic} consistency.
% (see \S\ref{sec:consistency} below).

\paragraph{Syntactic Subjects and Objects.}
In the active sentence \sent{The fairy charms the elf}, the phrase \sent{the fairy} is the syntactic subject and \textit{the elf} the syntactic object. In the passive \sent{The elf is charmed (by the fairy)}, the phrase \sent{the elf} is now the syntactic subject, even though in both sentences, the fairy is the charmer and the elf the charmee. 
% not in workshop submission 
Similarly, \textit{the fairy} is the syntactic subject in the intransitive sentence \sent{The fairy glows}.

\begin{figure}
    % transitive constant
    \begin{subfigure}[b]{0.32\linewidth}
        \centering
        \includegraphics[scale=\defAMRScale]{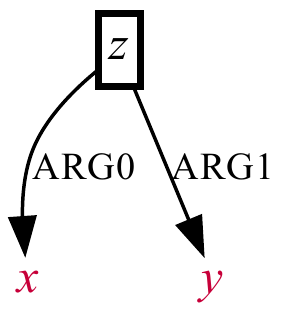}
        \caption{Transitive constants, with label $z$ and sources $x,y$}
        \label{fig:transitive}
    \end{subfigure}
    % active AM tree
    \begin{subfigure}[b]{0.67\linewidth}
    \centering
    \pgfkeys{/pgf/inner sep=0.1em} % words close together
    \begin{forest}  
    sn edges/.style={for tree={
    align=center, font=\small, l sep=1em, l=1em,
    parent anchor=south, child anchor=north}},  
    sn edges,
    [\G{charm}
        [,phantom
            [The, tier=word]
        ]
        [\G{fairy}, edge=->, edge label = {node [midway, left, xshift=-.5em] {\app{\subj}}}, l=2em
            [fairy,tier=word, edge=dashed]
        ]
        [charms, tier=word, edge=dashed]
        [,phantom
            [the, tier=word]
        ]
        [\G{elf}, edge=->, edge label = {node [midway, right, xshift=.7em] {\app{\obj}}}, l=2em
            [elf,tier=word, edge=dashed]
        ]
    ]
    \end{forest}
    \caption{Active sentence defines \subj{} and \obj{}}
    \label{fig:active}
    \end{subfigure}
        \begin{subfigure}[b]{0.23\linewidth}
        \centering
        \includegraphics[scale=\defAMRScale]{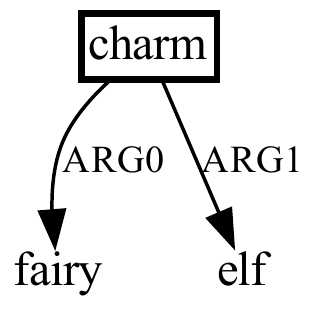}
        \caption{AMR for both sentences}
        \label{fig:amr-charm}
    \end{subfigure}
    % passive AM tree
    \begin{subfigure}[b]{0.71\linewidth}
    \centering
    \pgfkeys{/pgf/inner sep=0.1em} % words close together
    % syntactically consistent passive
    \begin{forest}  
    sn edges/.style={for tree={
    align=center, font=\small, l sep=0.5em, l=0.5em,
    parent anchor=south, child anchor=north}},  
    sn edges,
    [\G{charmP}
        [,phantom
            [The, tier=word]
        ]
        [\G{elf}, edge=->, edge label = {node [midway, left, xshift=-.8em] {\app{\subj}}}, l=2em
            [elf,tier=word, edge=dashed]
        ]
        [,phantom
            [is, tier=word]
        ]
        [charmed, tier=word, edge=dashed]
        [,phantom
            [by the, tier=word]
        ]
        [\G{fairy}, edge=->, edge label = {node [midway, right, xshift=.8em] {\app{\obj}}}, l=2em
            [fairy,tier=word, edge=dashed]
        ]
    ]
    \end{forest}
    \caption{Neural analysis of passive sentences mirrors surface syntax}
	\label{fig:passive-neural}
    \end{subfigure}
    % deep passive AM tree
    \begin{subfigure}[t]{\linewidth}
    \pgfkeys{/pgf/inner sep=0.1em} % words close together
    \centering
    % EM passive
    \begin{forest}  
    sn edges/.style={for tree={
    align=center, font=\small, l sep=0.5em, l=0.5em,
    parent anchor=south, child anchor=north}},  
    sn edges,
    [\G{charm}
        [,phantom
            [The, tier=word]
        ]
        [\G{elf}, edge=->, edge label = {node [midway, left, xshift=-.8em] {\app{\obj}}}, l=2em
            [elf,tier=word, edge=dashed]
        ]
        [,phantom
            [is, tier=word]
        ]
        [charmed, tier=word, edge=dashed]
        [,phantom
            [by the, tier=word]
        ]
        [\G{fairy}, edge=->, edge label = {node [midway, right, xshift=.8em] {\app{\subj}}}, l=2em
            [fairy,tier=word, edge=dashed]
        ]
    ]
    \end{forest}
    \caption{EM analysis of passives uses \app{\obj} for syntactic subject}
	\label{fig:passive-em}
    \end{subfigure}
    \caption{AMR examples of active and passive. See Fig.\ \ref{fig:as-graphs} for graph constants.}
    \label{fig:amr-examples}
\end{figure}

\paragraph{Intra-Phenomenon Consistency.}
For both the EM and neural method, we found completely consistent source allocations for active transitive verbs in all four sembanks. These source allocations were also the overwhelming favourite graph constants for two-argument predicates (72-92\%), and the most common sources used by Apply operations (94-98\%). 
For example, in AMR, the graph constant template in Fig.\ \ref{fig:transitive} appears 26,653 times in the neural parser output. 74\% of these used sources $x=$ \src{s$_1$} and $y=$ \src{s$_2$} (from $\sources{}=\left\{\src{s$_1$}, \src{s$_2$}, \src{s$_3$}\right\}$). All active transitive sentences in our sample used this source allocation, so we call this the \textit{active} graph constant (e.g.\ \G{charm} in Fig.\ \ref{fig:as-graphs}) and refer to the sources \src{s$_1$} and \src{s$_2$} as \subj{} and \obj{} respectively, for \textit{subject} and \textit{object}. 
All four sembanks showed this kind of consistency; when we refer to \subj{} and \obj{} sources below, we mean whichever two sources displayed the same behaviour as \src{s$_1$} and \src{s$_2$} in AMR.

% not in workshop submission
 All four graphbanks are also highly consistent in their modifiers: classical modifiers such as adjectives are nearly universally adjoined with one consistent source -- we refer to it as \modsrc{} -- and \modify{\modsrc{}} is the overwhelming favourite (90-99\%) for \modify{} operations.

%For example, in PAS the graph constant with two edges labelled \el{verb\_ARG1} and \el{verb\_ARG2} is used by the neural method 52,834 times, of which 38,094 have sources \src{s1} and \src{s0} respectively. We name these sources \subj{} and \obj{} respectively and call the graph constant the \textit{active} graph constant. Hand-analysis of a sample of the sentences using this graph constant found only active transitive verbs like that in Fig.\ \ref{fig:active}. 909 of the sentences used a graph constant with swapped sources, giving \subj{} to \el{verb\_ARG2} and \obj{} to \el{very\_ARG1}. We call this the \textit{passive} graph constant. All sampled sentences using the passive graph constant were passive sentences.\footnote{Most of the remaining were for control-like structures (as in \ref{fig:as-graphs}) and relative clauses. Relatives are modifiers, and we found a preference for consistency of operation (the vast majority of \modify{}s are \modify{\src{s2}}) over consistency of graph constant, so these use source \src{s2} for the relativised argument.} While other graphbanks don't annotate POS on their edge labels, and therefore have more diversity of uses for their active graph constants, all used active graph constants for active sentences and passive graph constants for passives consistently.

\paragraph{Cross-Phenomenon Consistency.}
%Consider Fig.\ \ref{fig:amr-examples} showing analyses of active and passive sentences. We call a \todo{grammar?} \textit{syntactically consistent} if it uses the analysis in \ref{fig:active} for actives and \ref{fig:passive-neural} for passives because in both sentences, the \textit{syntactic subject} fills an \app{\subj} slot, even though in the passive sentence the syntactic subject \textit{unicorn} is the patient/semantic object. 
We call a parser \textit{syntactically consistent} if its syntactic subjects fill the \subj{} slot, regardless of their semantic role. A syntactically consistent parser would acquire the AMR in Fig.~\ref{fig:amr-charm} from the active sentence by the analysis in Fig.~\ref{fig:active}, and from the passive sentence by the analysis in Fig.~\ref{fig:passive-neural}, with the passive constant \G{charmP} from Fig.~\ref{fig:as-graphs}. 

The neural parser is syntactically consistent: in all sembanks, it uses the same source \subj{} for syntactic subjects in passives as for actives. EM, conversely, prefers to  use the same graph constants for active and passives, flipping the \app{} edges to produce syntactically inconsistent trees as in Fig.~\ref{fig:passive-em}. Single-argument predicates are also syntactically consistent in the neural model, using \subj{} for subjects and \obj{} for objects, while EM picks one source.
%uses the passive graph constant only when forced to by reentrancies. For example, PAS auxilliaries, including \textit{be}, have an \subj{} source for the syntactic subject of the passive verb, and so in order to merge those nodes, PAS uses the passive graph constant for passives.
% not in workshop submission
The heuristics in \newcite{lindemann-etal-2019-compositional} have passive constants, but use them
% in the training data 
only when forced to, e.g.\ when coordinating active and passive.

% not in workshop submission
%When the verb has only one argument, the neural method, being syntactically consistent, chooses \obj{} for syntactic objects (e.g. objects of gerunds, imperatives) and \subj{} for syntactic subjects (e.g. subjects of intransitive verbs). EM consistently chooses the same source for both (74-99\%), using \subj{} for AMR, DM and PAS, and \obj{} for PSD. 
 
% {\small
% \ex. 
% \a. The cat slept. \hfill \textsl{Intransitive} \label{ex:intransitive}
% \b. The dog chased the cat. \hfill \textsl{Active} \label{ex:active}
% \b. The cat was chased (by the dog). \hfill \textsl{Passive} \label{ex:passive}

% }

% not in workshop submission
Finally, we compute the entropy of the graph constants for the best trees of the training set as $\sum_G f(g)\ln f(G)$, where $f(G)$ is the frequency of constant $G$ in the trees.%, i.e.~the count of $G$ divided by the total count of constants in the trees. 
The entropies are between 2 and 3 nats, but are consistently lower for EM than the neural method, by 0.031 to 0.079 nats. Considering that the neural method achieves higher parsing accuracies, using the most common graph constants and edges possible evidently is not always optimal for performance. The syntactic regularities exploited by the neural method may contribute to its improved performance.

%% file: sampling.tex
\section{Sampling Method for hand analysis}\label{app:sampling}

To sample trees, we compute for each sentence in the training set the best tree from its tree automaton, according to the neural weights of the best performing epoch. This ensures the AM trees evaluate to the correct graph. We then sample trees from this set for hand-analysis.

To get relevant sentences, we sampled 5-to-15-word sentences with graph constants from the following six categories: 

\paragraph{Transitive verbs:} graph constants with a labeled root and two arguments with edges labelled as  in Table \ref{tab:transitives}:

\begin{table}[htb]
    \centering
    \begin{tabular}{c|cc}
        
        Sembank & subject & object \\
        \hline
        AMR & \textsc{arg0} & \textsc{arg1} \\
        DM & \textsc{arg1} & \textsc{arg2} \\
        PAS & verb\_\textsc{arg1} & verb\_\textsc{arg2} \\
        PSD & ACT\_arg & PAT\_arg \\
    \end{tabular}
    \caption{Transitive verbs}
    \label{tab:transitives}
\end{table}

As explained in the main text, we define the \emph{active} constants as those with the most common source allocation, and the \textit{passive} constants as those with the active source allocation flipped. 
We sampled both active and passive source allocations.

\paragraph{Verbs with one argument:} Graph constants just like the transitive ones but lacking one of the arguments. There are four of these, given both source allocations.  

Generally these graph constants are used for more than just verbs; for each of the six categories we sampled until we had ten relevant sentences. We visualised the AM trees and categorised the phenomena, for example active or passive verbs, nominalised verbs, imperatives, relative clauses, gerund modifiers, and so forth.

To answer the question of whether the parser used consistent constants for active and passive transitive sentences, we sampled until we had ten sentences with active or passive main verbs. For the single-argument verbs, we also looked at nominalised verbs, modifiers, and so forth. (Sampling and visualisation scripts will be available together with the rest of our code on GitHub.)

%%% Local Variables:
%%% mode: latex
%%% TeX-master: "appendix"
%%% End:

%% file: appendix-all-trees-alg.tex
\section{An algorithm to obtain all AM dep-trees for a graph}\label{app:percolations}

Let $G$ be a graph partitioned into blobs. Let $\mathcal{U}_G$ be the set of unrolled graphs for $G$ that can be obtained by Algorithm~2 by varying the queue order.

Let further $\mathcal{M}_G$ be the set of results of Algorithm~\ref{alg:modswitch} below for every input AM dep-tree $T=C_U$ for $U\in\mathcal{U}_G$ and every choice of set $M$ as specified in the algorithm. Algorithm~\ref{alg:modswitch} switches the order of two nodes \nn{m} and \nn{k}, making \nn{k} the head of the subtree previously headed by \nn{m}. This change of head is only possible when the incoming edge of \nn{m} is labeled \modify{} (for \app{}, the change of head changes the evaluation result). It also requires a \modify{} edge between \nn{m} and \nn{k}; an \app{} edge with this type of swap would lead to a non-well-typed graph.

Finally, let  $\mathcal{R}_G$ be the set of results of Algorithm~\ref{alg:percolExtended} for every input AM dep-tree $T\in\mathcal{M}_G$ and any valid choice of $R$ and $\text{RT}$ (valid as described in the algorithm). Algorithm~\ref{alg:percolExtended} is like Algorithm~1 for reentrancy resolution, but can have resolution targets \resolutionTarget{n} that are higher in the tree than the lowest common ancestor of \nn{n} and the \refnode{n} nodes. Further, Algorithm~\ref{alg:percolExtended} uses the same methodology to also move nodes that do not need resolution to become descendents of a \sortof{resolution target} higher in the tree (i.e.~$R$ here can now also contain nodes for which no REF node exists).

Then the following Theorem~\ref{thm:correctComplete} holds:

\begin{theorem}\label{thm:correctComplete}
Let $G$ be a graph partitioned into blobs, and let $\mathcal{T}_G$ be the set of all well-typed AM dep-trees with placeholder sources, using that blob partition, that evaluate to $G$. Then if $\mathcal{T}_G=\emptyset$, all AM dep-trees in $\mathcal{R}_G$ are either not well-typed or do not evaluate to $G$. If however $\mathcal{T}_G\neq\emptyset$, then  $\mathcal{R}_G=\mathcal{T}_G$.
\end{theorem}

% more Pythony
\SetStartEndCondition{ }{}{}%
\SetKwProg{Fn}{def}{\string:}{}\SetKwFunction{Range}{range}%
\SetKw{KwTo}{in}\SetKwFor{For}{for}{\string:}{}%
\SetKwIF{If}{ElseIf}{Else}{if}{:}{elif}{else:}{}%
\SetKwFor{While}{while}{:}{fintq}%
\setcounter{algocf}{2}

\begin{algorithm}[t]
%\KwResult{an AM dep-tree or some unresolvable mess}
 Input: an AM dep-tree $T$ and a set $M$ of pairs of consecutive edges in $T$ of the form $\langle\edge{n}{m}{\modify{\ps{n}}}, \edge{m}{k}{\modify{\ps{m}}}\rangle$ such that no edge appears in multiple pairs.
 
  \For{$\langle\edge{n}{m}{\modify{\ps{n}}}, \edge{m}{k}{\modify{\ps{m}}}\rangle\in M$}{
  Replace \edge{n}{m}{\modify{\ps{n}}} in $T$ with \edge{n}{k}{\modify{\ps{n}}}\;
  Replace \edge{m}{k}{\modify{\ps{m}}} in $T$ with \edge{k}{m}{\app{\ps{m}}}\;
  Add $\termtype{m}$ (which always includes \ps{n}) to the request at \ps{m} in \lexicaltype{\nn{k}}\;
 }
 \Return{$T$}
 \caption{Modify-edge swapping}
 \label{alg:modswitch}
\end{algorithm}

\begin{algorithm}[t]
%\KwResult{an AM dep-tree or some unresolvable mess}
 Input: an AM dep-tree $T$; a set $R\supseteq \{n\in N_G ~|~ \exists\ \text{\refnode{n} node in $T$}\}$; and a map $\text{RT}$ that assigns to each node $n\in R$ a resolution target \resolutionTarget{n}, that is at least as high as the lowest common ancestor of \nn{n} and all \refnode{n} nodes (if they exist), and that satisfies the conditions of Theorem~1.
 
 \While{$R\neq \emptyset$}{
  Pick a $y\in R$ s.t. there is no $x\in R$, $x\neq y$, with $y$ on an $x$-resolution path\;
  \For{$p \in y$-resolution paths}{
  \For{ $\edge{n}{m}{\app{}} \in p$\label{algline:percol:edgeloop}}{
  \eIf{\text{\nn{m} is \nn{y} or labeled \refnode{y}}}{
   Add $\termtype{y}$ to the request at \ps{y} in \lexicaltype{$n$}\;\label{algline:percol:bottomedge}
   }{
   Add \ps{y}[$\termtype{y}$] to the request at \ps{m} in \lexicaltype{$n$}\label{algline:percol:middleedge}\;
  }
  }}
  Move the subtree of $T$ rooted at $y$ up to be an \app{\ps{y}} daughter of \resolutionTarget{y}, unless $\resolutionTarget{y}=y$\label{algline:percol:moveEdge}\;
  Delete all \refnode{y} nodes from $T$\label{algline:percol:delREF}\;
  $R \gets R -\{y\}$
 }
 \Return{$T$}
 \caption{Extended reentrancy resolution}
 \label{alg:percolExtended}
\end{algorithm}

%% file: other-details.tex
\section{Additional Details}

\begin{itemize}
\item AMR F-scores are \textit{Smatch scores} \citep{cai-knight-2013-smatch}
\item DM, PAS and PSD: We compute labeled F-score with  the  evaluation  toolkit  that  was  developed for the SDP shared task:  \url{https://github.com/semantic-dependency-parsing/toolkit}
\item We use the standard train/dev/test split for all corpora
\item AMR corpus available through \url{https://amr.isi.edu/download.html} (requires LDC license)
\item SDP corpora available through \url{https://catalog.ldc.upenn.edu/LDC2016T10} (requires LDC license)
\end{itemize}

 \paragraph{Number of source names.}
 We experimented with different numbers of source names in the joint neural method (Table~\ref{tab:sources}). Mostly, three source names were most effective, except for PSD, where four were most effective. Two source names are not enough to model many common phenomena (e.g.~ditransitive verbs, coordination of verbs); graphs containing these phenomena cannot be decomposed with two sources and are removed from the training set, reducing parsing accuracy.
 The higher performance of PSD with four sources may stem from PSD using flat coordination structures which require more source names; although this is also true for AMR where four source names are not beneficial.
The drop with six source names may come from the fact that the latent space grows rapidly with more sources, making it harder to learn consistent source assignments.

\paragraph{Hyperparameters.}
See Table~\ref{tab:hyper}.

\begin{table}[t]
\centering
\small
\begin{tabular}{rrrrr}
\multicolumn{1}{l}{\# sources} & DM            & PAS           & PSD           & AMR                   \\ 
\hline
2                              & 92.2          & 91.9          & 75.6          & 74.3                  \\
3                              & \textbf{94.5} & \textbf{94.8} & 82.7          & \textbf{76.5}         \\
4                              & 94.4          & 94.7          & \textbf{83.4} & 75.9                  \\
6                              & 92.3          & 93.6          & 80.1          & 73.4  
\end{tabular}
\vspace{-5pt}
\caption{Development set accuracies of the neural method for different numbers of source names.}\label{tab:sources}
\end{table}

\begin{table}
	\small
	\begin{tabular}{ll}
		\toprule
		Activation function & tanh \\
		Optimizer & Adam \\
		Learning rate & 0.001 \\
		Epochs & 100 \\
		\midrule
		Dim of lemma embeddings & 64 \\
		Dim of POS embeddings & 32 \\
		Dim of NE embeddings & 16 \\
		Minimum lemma frequency & 7 \\
		\midrule
		Hidden layers in all MLPs & 1 \\
		\midrule
		Hidden units in LSTM (per direction) & 256 \\
		Hidden units in edge existence MLP & 256 \\
		Hidden units in edge label MLP & 256 \\
		Hidden units in supertagger MLP & 1024 \\
		Hidden units in lexical label tagger MLP & 1024 \\
		\midrule
		Layer dropout in LSTMs & 0.3 \\
		Recurrent dropout in LSTMs & 0.4 \\
		Input dropout & 0.3\\
		Dropout in edge existence MLP & 0.0 \\
		Dropout in edge label MLP & 0.0 \\
		Dropout in supertagger MLP & 0.4 \\
		Dropout in lexical label tagger MLP & 0.4 \\
		\bottomrule
	\end{tabular}
	\caption{Common hyperparameters used in all experiments (the random trees, random weights and EM weights baselines use 40 epochs since they converge faster). For a complete description of the neural architecture, see \newcite{lindemann-etal-2019-compositional} and its supplementary materials.}
	\label{tab:hyper}
\end{table}

%%% Local Variables:
%%% mode: latex
%%% TeX-master: "appendix"
%%% End:

%% file: unsupervised2020.bbl
\begin{thebibliography}{29}
\expandafter\ifx\csname natexlab\endcsname\relax\def\natexlab#1{#1}\fi

\bibitem[{Anikina et~al.(2020)Anikina, Koller, and
  Roth}]{anikina-etal-2020-predicting}
Tatiana Anikina, Alexander Koller, and Michael Roth. 2020.
\newblock \href {https://www.aclweb.org/anthology/2020.crac-1.4} {Predicting
  coreference in {A}bstract {M}eaning {R}epresentations}.
\newblock In \emph{Proceedings of the Third Workshop on Computational Models of
  Reference, Anaphora and Coreference}, pages 33--38, Barcelona, Spain
  (online). Association for Computational Linguistics.

\bibitem[{Artzi et~al.(2015)Artzi, Lee, and Zettlemoyer}]{ArtziLZ15}
Yoav Artzi, Kenton Lee, and Luke Zettlemoyer. 2015.
\newblock \href {https://www.aclweb.org/anthology/D15-1198/} {Broad-coverage
  {CCG} {Semantic} {Parsing} with {AMR}}.
\newblock In \emph{Proceedings of the 2015 Conference on Empirical Methods in
  Natural Language Processing}.

\bibitem[{Baker(1979)}]{baker-io}
J.~K. Baker. 1979.
\newblock \href {https://doi.org/10.1121/1.2017061} {Trainable grammars for
  speech recognition}.
\newblock \emph{The Journal of the Acoustical Society of America},
  65(S1):S132--S132.

\bibitem[{Banarescu et~al.(2013)Banarescu, Bonial, Cai, Georgescu, Griffitt,
  Hermjakob, Knight, Koehn, Palmer, and Schneider}]{amBanarescuBCGGHKKPS13}
Laura Banarescu, Claire Bonial, Shu Cai, Madalina Georgescu, Kira Griffitt, Ulf
  Hermjakob, Kevin Knight, Philipp Koehn, Martha Palmer, and Nathan Schneider.
  2013.
\newblock \href {http://aclweb.org/anthology/W13-2322} {{A}bstract {M}eaning
  {R}epresentation for {Sembanking}}.
\newblock In \emph{Proceedings of the 7th Linguistic Annotation Workshop and
  Interoperability with Discourse}.

\bibitem[{Bevilacqua et~al.(2021)Bevilacqua, Blloshmi, and
  Navigli}]{bevilacqua2021one}
Michele Bevilacqua, Rexhina Blloshmi, and Roberto Navigli. 2021.
\newblock \href
  {https://github.com/SapienzaNLP/spring/blob/main/docs/preprint.pdf} {One
  {SPRING} to rule them both: Symmetric {AMR} semantic parsing and generation
  without a complex pipeline}.
\newblock In \emph{Proceedings of AAAI}.

\bibitem[{Cai and Knight(2013)}]{cai-knight-2013-smatch}
Shu Cai and Kevin Knight. 2013.
\newblock \href {https://www.aclweb.org/anthology/P13-2131} {{S}match: an
  evaluation metric for semantic feature structures}.
\newblock In \emph{Proceedings of the 51st Annual Meeting of the Association
  for Computational Linguistics (Volume 2: Short Papers)}, pages 748--752,
  Sofia, Bulgaria. Association for Computational Linguistics.

\bibitem[{Chen et~al.(2018)Chen, Sun, and Wan}]{chen-etal-2018-accurate}
Yufei Chen, Weiwei Sun, and Xiaojun Wan. 2018.
\newblock \href {https://www.aclweb.org/anthology/P18-1038} {Accurate
  {SHRG}-based semantic parsing}.
\newblock In \emph{Proceedings of the 56th Annual Meeting of the Association
  for Computational Linguistics (Volume 1: Long Papers)}, pages 408--418,
  Melbourne, Australia. Association for Computational Linguistics.

\bibitem[{Comon et~al.(2007)Comon, Dauchet, Gilleron, Jacquemard, Lugiez,
  Tison, Tommasi, and L\"{o}ding}]{ComonDGJLTL07}
Hubert Comon, Max Dauchet, R\'{e}mi Gilleron, Florent Jacquemard, Denis Lugiez,
  Sophie Tison, Marc Tommasi, and Christof L\"{o}ding. 2007.
\newblock \href {http://tata.gforge.inria.fr/} {\emph{Tree Automata techniques
  and applications}}.
\newblock published online - \url{http://tata.gforge.inria.fr/}.

\bibitem[{Copestake and Flickinger(2000)}]{copestake00:_englis_hpsg}
Ann Copestake and Dan Flickinger. 2000.
\newblock \href {https://www.aclweb.org/anthology/L00-1276/} {An open-source
  grammar development environment and broad-coverage english grammar using
  {HPSG}}.
\newblock In \emph{Proceedings of the Second conference on Language Resources
  and Evaluation (LREC)}.

\bibitem[{Copestake et~al.(2001)Copestake, Lascarides, and
  Flickinger}]{copestake2001algebra}
Ann Copestake, Alex Lascarides, and Dan Flickinger. 2001.
\newblock \href {https://www.aclweb.org/anthology/P01-1019/} {An algebra for
  semantic construction in constraint-based grammars}.
\newblock In \emph{Proceedings of the 39th ACL}.

\bibitem[{Courcelle and Engelfriet(2012)}]{CourcelleE12}
Bruno Courcelle and Joost Engelfriet. 2012.
\newblock \href {https://www.labri.fr/perso/courcell/Book/TheBook.pdf}
  {\emph{Graph Structure and Monadic Second-Order Logic, a Language Theoretic
  Approach}}.
\newblock Cambridge University Press.

\bibitem[{Devlin et~al.(2019)Devlin, Chang, Lee, and Toutanova}]{BERT}
Jacob Devlin, Ming-Wei Chang, Kenton Lee, and Kristina Toutanova. 2019.
\newblock \href {https://www.aclweb.org/anthology/N19-1423} {{BERT}:
  Pre-training of deep bidirectional transformers for language understanding}.
\newblock In \emph{Proceedings of the 2019 Conference of the North {A}merican
  Chapter of the Association for Computational Linguistics: Human Language
  Technologies, Volume 1 (Long and Short Papers)}, pages 4171--4186,
  Minneapolis, Minnesota. Association for Computational Linguistics.

\bibitem[{Eisner(2016)}]{eisner2016inside}
Jason Eisner. 2016.
\newblock \href {https://www.aclweb.org/anthology/W16-5901/} {Inside-outside
  and forward-backward algorithms are just backprop (tutorial paper)}.
\newblock In \emph{Proceedings of the Workshop on Structured Prediction for
  NLP}, pages 1--17.

\bibitem[{Fancellu et~al.(2019)Fancellu, Gilroy, Lopez, and
  Lapata}]{fancellu-etal-2019-semantic}
Federico Fancellu, Sorcha Gilroy, Adam Lopez, and Mirella Lapata. 2019.
\newblock \href {https://doi.org/10.18653/v1/D19-1278} {Semantic graph parsing
  with recurrent neural network {DAG} grammars}.
\newblock In \emph{Proceedings of the 2019 Conference on Empirical Methods in
  Natural Language Processing and the 9th International Joint Conference on
  Natural Language Processing (EMNLP-IJCNLP)}, pages 2769--2778, Hong Kong,
  China. Association for Computational Linguistics.

\bibitem[{Fern{\'a}ndez-Gonz{\'a}lez and
  G{\'o}mez-Rodr{\'\i}guez(2020)}]{lys-2020-transition}
Daniel Fern{\'a}ndez-Gonz{\'a}lez and Carlos G{\'o}mez-Rodr{\'\i}guez. 2020.
\newblock \href {https://doi.org/10.18653/v1/2020.acl-main.629}
  {Transition-based semantic dependency parsing with pointer networks}.
\newblock In \emph{Proceedings of the 58th Annual Meeting of the Association
  for Computational Linguistics}, pages 7035--7046, Online. Association for
  Computational Linguistics.

\bibitem[{Flickinger et~al.(2017)Flickinger, Haji{\v c}, Ivanova, Kuhlmann,
  Miyao, Oepen, and Zeman}]{OpenSDP}
Dan Flickinger, Jan Haji{\v c}, Angelina Ivanova, Marco Kuhlmann, Yusuke Miyao,
  Stephan Oepen, and Daniel Zeman. 2017.
\newblock \href {http://hdl.handle.net/11234/1-1956} {Open {SDP} 1.2}.
\newblock {LINDAT}/{CLARIN} digital library at the Institute of Formal and
  Applied Linguistics ({{\'U}FAL}), Faculty of Mathematics and Physics, Charles
  University.

\bibitem[{Groschwitz(2019)}]{GroschwitzDiss}
Jonas Groschwitz. 2019.
\newblock \href {http://www.coli.uni-saarland.de/~jonasg/thesis.pdf}
  {\emph{Methods for taking semantic graphs apart and putting them back
  together again}}.
\newblock Ph.D. thesis, Macquarie University and Saarland University.

\bibitem[{Groschwitz et~al.(2017)Groschwitz, Fowlie, Johnson, and
  Koller}]{groschwitz-etal-2017-constrained}
Jonas Groschwitz, Meaghan Fowlie, Mark Johnson, and Alexander Koller. 2017.
\newblock \href {https://www.aclweb.org/anthology/W17-6810} {A constrained
  graph algebra for semantic parsing with {AMR}s}.
\newblock In \emph{{IWCS} 2017 - 12th International Conference on Computational
  Semantics - Long papers}.

\bibitem[{Groschwitz et~al.(2018)Groschwitz, Lindemann, Fowlie, Johnson, and
  Koller}]{groschwitz18:_amr_depen_parsin_typed_seman_algeb}
Jonas Groschwitz, Matthias Lindemann, Meaghan Fowlie, Mark Johnson, and
  Alexander Koller. 2018.
\newblock \href {http://aclweb.org/anthology/P18-1170} {A{MR} {Dependency
  Parsing with a Typed Semantic Algebra}}.
\newblock In \emph{Proceedings of ACL}.

\bibitem[{Havrylov et~al.(2019)Havrylov, Kruszewski, and
  Joulin}]{havrylov-etal-2019-cooperative}
Serhii Havrylov, Germ{\'a}n Kruszewski, and Armand Joulin. 2019.
\newblock \href {https://doi.org/10.18653/v1/N19-1115} {Cooperative learning of
  disjoint syntax and semantics}.
\newblock In \emph{Proceedings of the 2019 Conference of the North {A}merican
  Chapter of the Association for Computational Linguistics: Human Language
  Technologies, Volume 1 (Long and Short Papers)}, pages 1118--1128,
  Minneapolis, Minnesota. Association for Computational Linguistics.

\bibitem[{He and Choi(2020)}]{bertbaseline}
Han He and Jinho Choi. 2020.
\newblock \href {https://arxiv.org/abs/1908.04943} {Establishing strong
  baselines for the new decade: Sequence tagging, syntactic and semantic
  parsing with {BERT}}.
\newblock In \emph{The Thirty-Third International Flairs Conference}.

\bibitem[{Herzig and Berant(2020)}]{herzig2020spanbased}
Jonathan Herzig and Jonathan Berant. 2020.
\newblock \href {http://arxiv.org/abs/2009.06040} {Span-based semantic parsing
  for compositional generalization}.

\bibitem[{Kitaev and Klein(2018)}]{kitaev-klein-2018-constituency}
Nikita Kitaev and Dan Klein. 2018.
\newblock \href {https://doi.org/10.18653/v1/P18-1249} {Constituency parsing
  with a self-attentive encoder}.
\newblock In \emph{Proceedings of the 56th Annual Meeting of the Association
  for Computational Linguistics (Volume 1: Long Papers)}, pages 2676--2686,
  Melbourne, Australia. Association for Computational Linguistics.

\bibitem[{Lindemann et~al.(2019)Lindemann, Groschwitz, and
  Koller}]{lindemann-etal-2019-compositional}
Matthias Lindemann, Jonas Groschwitz, and Alexander Koller. 2019.
\newblock \href {https://doi.org/10.18653/v1/P19-1450} {Compositional semantic
  parsing across graphbanks}.
\newblock In \emph{Proceedings of the 57th Annual Meeting of the Association
  for Computational Linguistics}, pages 4576--4585, Florence, Italy.
  Association for Computational Linguistics.

\bibitem[{Lindemann et~al.(2020)Lindemann, Groschwitz, and
  Koller}]{lindemann-etal-2020-fast}
Matthias Lindemann, Jonas Groschwitz, and Alexander Koller. 2020.
\newblock \href {https://doi.org/10.18653/v1/2020.emnlp-main.323} {Fast
  semantic parsing with well-typedness guarantees}.
\newblock In \emph{Proceedings of the 2020 Conference on Empirical Methods in
  Natural Language Processing (EMNLP)}, pages 3929--3951, Online. Association
  for Computational Linguistics.

\bibitem[{Maillard et~al.(2019)Maillard, Clark, and
  Yogatama}]{maillard2019jointly}
Jean Maillard, Stephen Clark, and Dani Yogatama. 2019.
\newblock \href {https://arxiv.org/abs/1705.09189} {Jointly learning sentence
  embeddings and syntax with unsupervised tree-{LSTMs}}.
\newblock \emph{Natural Language Engineering}, 25(4).

\bibitem[{Oepen et~al.(2015)Oepen, Kuhlmann, Miyao, Zeman, Cinkov\'{a},
  Flickinger, Haji\v{c}, and Ure\v{s}ov\'{a}}]{OepenKMZCFHU15}
Stephan Oepen, Marco Kuhlmann, Yusuke Miyao, Daniel Zeman, Silvie Cinkov\'{a},
  Dan Flickinger, Jan Haji\v{c}, and Zde\v{n}ka Ure\v{s}ov\'{a}. 2015.
\newblock \href {http://aclweb.org/anthology/S15-2153} {Semeval 2015 task 18:
  Broad-coverage semantic dependency parsing}.
\newblock In \emph{Proceedings of the 9th International Workshop on Semantic
  Evaluation (SemEval 2015)}.

\bibitem[{Peng et~al.(2015)Peng, Song, and Gildea}]{PengSG15}
Xiaochang Peng, Linfeng Song, and Daniel Gildea. 2015.
\newblock \href {https://www.aclweb.org/anthology/K15-1004/} {A synchronous
  hyperedge replacement grammar based approach for {AMR} parsing}.
\newblock In \emph{Proceedings of the 19th Conference on Computational Language
  Learning}.

\bibitem[{Vaswani et~al.(2017)Vaswani, Shazeer, Parmar, Uszkoreit, Jones,
  Gomez, Kaiser, and Polosukhin}]{vaswani2017attention}
Ashish Vaswani, Noam Shazeer, Niki Parmar, Jakob Uszkoreit, Llion Jones,
  Aidan~N. Gomez, undefinedukasz Kaiser, and Illia Polosukhin. 2017.
\newblock \href {https://research.google/pubs/pub46201/} {Attention is all you
  need}.
\newblock In \emph{Proceedings of the 31st International Conference on Neural
  Information Processing Systems}, NIPS'17, page 6000–6010, Red Hook, NY,
  USA. Curran Associates Inc.

\end{thebibliography}
